\documentclass{article}

\usepackage{arxiv}

\usepackage[utf8]{inputenc} % allow utf-8 input
\usepackage[T1]{fontenc}    % use 8-bit T1 fonts
\usepackage{hyperref}       % hyperlinks
\usepackage{url}            % simple URL typesetting
\usepackage{booktabs}       % professional-quality tables
\usepackage{amsfonts}       % blackboard math symbols
\usepackage{nicefrac}       % compact symbols for 1/2, etc.
\usepackage{microtype}      % microtypography
\usepackage{lipsum}
\usepackage{svg}
\usepackage{graphicx}
\graphicspath{ {./images/} }
\usepackage{amsmath} 
\usepackage{pdflscape}
\usepackage{xcolor}
\usepackage[normalem]{ulem}
\usepackage{comment}
 \usepackage{longtable}
\usepackage{tabularx}
\usepackage{caption} % for more complex captioning
\usepackage{subcaption}
\usepackage{longtable}
\usepackage{booktabs}
\usepackage[utf8]{inputenc}

\title{YOLOv1 to YOLOv10: A comprehensive review of YOLO variants and their application in the agricultural domain}

\author{
  \textbf{Mujadded Al Rabbani Alif}\textsuperscript{*} and \textbf{Muhammad Hussain}\\[1ex] % Adds a little space between names and affiliation
  \begin{minipage}[t]{0.90\textwidth}
    \scriptsize Department of Computer Science, Huddersfield University, Queensgate, Huddersfield HD1 3DH, UK; \\
    \textsuperscript{*}Correspondence: M.Alif@hud.ac.uk;
  \end{minipage}
}
\begin{document}
\maketitle
% Abstract (Do not insert blank lines, i.e. \\) 
\begin{abstract}This survey investigates the transformative potential of various YOLO variants, from YOLOv1 to the state-of-the-art YOLOv10, in the context of agricultural advancements. The primary objective is to elucidate how these cutting-edge object detection models can re-energise and optimize diverse aspects of agriculture, ranging from crop monitoring to livestock management. It aims to achieve key objectives, including the identification of contemporary challenges in agriculture, a detailed assessment of YOLO's incremental advancements, and an exploration of its specific applications in agriculture. This is one of the first surveys to include the latest YOLOv10, offering a fresh perspective on its implications for precision farming and sustainable agricultural practices in the era of Artificial Intelligence and automation. Further, the survey undertakes a critical analysis of YOLO's performance, synthesizes existing research, and projects future trends. By scrutinizing the unique capabilities packed in YOLO variants and their real-world applications, this survey provides valuable insights into the evolving relationship between YOLO variants and agriculture. The findings contribute towards a nuanced understanding of the potential for precision farming and sustainable agricultural practices, marking a significant step forward in the integration of advanced object detection technologies within the agricultural sector.
\end{abstract}

% Keywords

\section{Introduction}

In recent years, the intersection of computer vision and agriculture has experienced remarkable strides, unlocking a transformative era in precision farming and agricultural management~\cite{ariana2006near}. Among the key technologies driving this paradigm shift is the evolution of the You Only Look Once (YOLO) algorithm, a family of object detectors that have manifested exceptional efficiency and accuracy. This review aims to delve into the mainstream YOLO variants, starting with YOLOv1 and continuing with the latest advancements in the form of YOLOv10. In particular, this exploration seeks to unravel the emerging potential of YOLO variants in revolutionising agricultural practices and fostering sustainable advancements. The YOLO family of object detectors, introduced by Joseph Redmon in 2015 with YOLOv1, marked a watershed moment in the object detection catalogue of architectures. YOLO's distinctive characteristic lies in its ability to perform real-time object detection by dividing the input image into a grid matrix and predicting bounding boxes and class probabilities simultaneously~\cite{hussain2023yolo}. This shift from conventional two-stage methodologies significantly enhanced speed and maintained competitive accuracy, setting the stage for subsequent YOLO iterations. As YOLO progressed through its reforms, each variant addressed limitations in addition to introducing novel techniques for refining edge-based performance. The transition from YOLOv1 to YOLOv10 witnessed advancements in several fields, including architectural design, training strategies, and optimisation techniques. The later variants of YOLO are designed to tackle challenges such as small object targets, occlusions, and improving performance across diverse datasets. It is important to understand these complexities to fully appreciate the potential applicability of YOLO variants in complex sub-domains such as agriculture. 

Agriculture, being a multifaceted domain, demands robust and highly efficient tools for monitoring and management of crops, livestock, and environmental conditions~\cite{zhu2019rapid}. The integration of YOLO variants in agricultural applications is promising for revolutionising tasks such as crop monitoring~\cite{zaji2023wheat}, disease detection ~\cite{mao2010confirmation}, yield estimation~\cite{wang2018field}, and livestock management. The real-time capacity offered by YOLO, coupled with its accuracy and adaptability, make YOLO variants an attractive solution for addressing the evolving challenges in modern agriculture. Given the rising demands within the agricultural sector, computer vision stands out as a transformative force for several reasons:
            \begin{itemize}
                \item \textbf{Scale and Precision:} Automation, enabled by computer vision, accelerates large-scale and precise operations. Computer vision algorithms can facilitate the rapid and accurate analysis of visual data, allowing for the monitoring of vast agricultural landscapes with unprecedented levels of speed and precision.
                
                \item \textbf{Efficiency and Resource Optimisation:} The integration of computer vision can improve the efficiency of automated processes. Advanced image recognition and on-device analysis can enable the allocation of resources such as water, fertilizers, and land. This not only maximises yield but also promotes sustainable farming practices.
                
                \item \textbf{Real-time Response:} Computer vision contributes to real-time monitoring and analysis. Automated systems, fuelled by computer vision, can swiftly detect and respond to emerging challenges, such as disease outbreaks or pest invasions, ensuring a rapid and targeted intervention to alleviate potential losses and maintain crop health.
                
                \item \textbf{Data-Driven Decision-Making:} Computer vision augments automation by providing a wealth of real-time visual data. This data-driven approach can enable farmers and stakeholders to make informed decisions, improving overall farm management and strategic planning.
            \end{itemize}

In the subsequent sections of this review, we will navigate through key advancements of each YOLO variant, elucidating enhancements introduced in each variant. Subsequently, we will examine specific applications and the potential impact of YOLO variants in agriculture, investigating how these variants can contribute to sustainable farming practices and agricultural advancements. As we navigate through this review, it becomes clear that the fusion of cutting-edge computer vision technologies, represented by YOLO variants, with the agricultural landscape holds the promise of driving a new era of precision farming and resource optimization.
%%%%%%%%%%%%%%%%%%%%%%%%%%%%%%%%%%%%%%%%%%
\subsection{Survey Objective} 

This review endeavours to examine the transformative potential of YOLO variants, spanning from YOLOv1 to the state-of-the-art YOLOv10, in the realm of agricultural advancements. The overarching aim is to elucidate how these state-of-the-art architectures belonging to the YOLO family can reshape and optimise various facets of agriculture, ranging from crop monitoring to livestock management. The primary focus will be on harnessing the unique capabilities of YOLO variants to accommodate the dynamic challenges faced by the agricultural sector. Specifically, this review aims to achieve the following objectives:

\textbf{Assessment of YOLO Evolution:} We trace the fundamental advancements of each YOLO variant, examining the architectural enhancements, algorithmic refinements, and methodological innovations. By comprehensively understanding the incremental evolution of YOLO variants, we can affirm the technological strides that contribute to YOLO's applicability in diverse agricultural scenarios.

\textbf{Exploration of YOLO Applications in Agriculture:} We then explore the specific applications of YOLO variants in agricultural sub-domains. This involves the investigation of real-world use cases where YOLO has demonstrated high efficacy, such as crop monitoring, livestock tracking, and detection of anomalies in the agricultural landscape. By identifying these applications, we can gauge the versatility of YOLO in tackling the multifaceted challenges of modern agriculture.

\textbf{Critical Analysis of YOLO's Performance:} A critical review of the performance of YOLO variants in agricultural contexts is presented. This includes assessing metrics such as detection accuracy, processing speed, and adaptability to diverse agricultural environments. Through a nuanced analysis, we are able to ascertain the strengths and limitations of YOLO in meeting the specific demands of agriculture.

\textbf{Synthesis of Existing Research:} The survey aims to synthesize and analyse existing research studies that have explored the intersection of YOLO variants concerning the agricultural domain. By consolidating the findings of these works, we unlock key insights and discern common trends, paving the way for a comprehensive understanding of the current landscape.

\textbf{Projection of Future Trends:} We anticipate future trends crucial for benchmarking the impact of YOLO variants in agriculture. This objective involves extrapolating from current research and technological trajectories to envisaging potential advancements, challenges, and emerging applications of YOLO in the agricultural domain.

In brief, this review aspires to present a holistic manifestation of the evolving relationship between YOLO variants and agriculture, unravelling the layers of innovation that hold the promise of ushering in a new era of precision farming and sustainable agricultural practices.

\subsection{Organization of Paper}
This review initiates by delineating agricultural challenges, followed by an introduction to Convolutional Neural Networks (CNNs) to furnish readers with foundational insights into the principles underpinning the YOLO framework. Subsequently, a general introduction to object detection techniques is presented to establish the necessary contextual background for the subsequent discourse.

Furthermore, the review undertakes a comprehensive examination of the evolutionary trajectories of YOLO architectures, systematically investigating the modifications and enhancements introduced by each variant.

Subsequent to this examination, the review scrutinizes the application of YOLO variants in sub-domains of agriculture, encompassing crop diseases, pest infestations, resource optimisation, and precision farming. In the conclusive segments, the findings are summarised in detail, culminating in a comprehensive assessment of YOLO variants as a transformative solution for agricultural domains.

\section{Convolutional Neural Networks (CNN)}

Deep Learning (DL) has emerged as a multi-domain innovation amid the popularity of various Machine Learning (ML) techniques like Decision Trees (DT), Support Vector Machines (SVM), KMeans, Multilayer Perceptron (MLP), and Artificial Neural Networks (ANN). DL, a subset of ML and an Artificial Intelligence (AI) component, has demonstrated remarkable success across diverse domains. Its applications include biological data handling~\cite{20}, speech recognition~\cite{21}, character recognition~\cite{22}, micro-blogs~\cite{23}, text classification~\cite{24}, unstructured text data mining with fault classification~\cite{25}, gene expression~\cite{26}, bolt detection~\cite{boltvision, alif2024lightweight}, pallet damage detection~\cite{attention}, automatic landslide detection~\cite{27}, video processing, including caption generation~\cite{28}, intrusion detection~\cite{29} and stock market prediction~\cite{30}. However, these examples only scratch the surface of Deep Learning’s vast potential.

In the context of this review, computer vision involves training machines to comprehend and interpret visual content at a sophisticated level. This field includes various subfields such as object detection~\cite{31}, image restoration, scene or object recognition, pose and motion estimation, object segmentation, and video tracking, among others~\cite{32}. Unlike conventional image processing, which requires manual feature extraction through the definition of feature descriptors, deep learning architectures serve as automatic feature extractors. This makes deep learning a lucrative alternative, enabling researchers to overcome conventional image processing constraints and focus more on improving application-specific performance.

Deep learning models encompass various techniques, including Recurrent Neural Networks (RNNs) for sequential data processing and their architectural variants, such as Long Short-Term Memory (LSTM) and Gated Recurrent Unit (GRU) for memory and context preservation~\cite{16}. Convolutional Neural Networks (CNNs) specialize in visual perception tasks involving image data, as other DL algorithms like ANNs struggle with scaling inefficiencies when confronted with high-dimensional input data, such as images.

The architectural footprint of Convolutional Neural Networks (CNNs) at an abstract level consists of a set of convolution, pooling, and activation functions that transform inputs through a staged process to reach the appropriate output.

A fundamental component within convolutional blocks of a CNN is defining the number of kernels/filters and their respective dimensions. These are crucial for feature extraction, providing low-level spatial information to subsequent layers for developing semantic relationships~\cite{33}.

\begin{equation}
q_i^l = f\left(b_i + \sum_{j=0}^{d-1} w_{i+j} x_{i+j}\right)
\end{equation}

\begin{equation}
q_{ij}^l = f\left(b_{ij} + \sum_{k=0}^{d1-1}\sum_{l=0}^{d2-1} w_{(i+k)(j+l)}x_{(i+k)(j+l)}\right)
\end{equation}

Similar to ANNs, the weights (\(w\)) and spatial input (\(x\)) are multiplied via the dot product. After introducing the bias term (\(b\)), a non-linear activation function (\(f\)) is applied.

Equations (1) and (2) express the convolution operation for the input, respectively. In these equations, \(q_i^l\) represents the output of the \(i\)-th neuron in layer \(l\). For text input, the filter size is denoted as \(d\), while for visual input, \(d_1\) and \(d_2\) represent the filter width and filter height.

The output post-convolving is subjected to pooling to extract prominent features through aggregation, i.e., downsampling along the spatial dimensions.

Multifaceted aggregation, i.e., pooling frameworks, are available~\cite{34}, such as average pooling, sum pooling, and max pooling. For example, the mathematical composition of the latter, i.e., the max pooling function, is expressed by equation 3.

\begin{equation}
q_i^l = \max(q_{(i-j)}^{(l-1)}, q_{(i+j)}^{(l-1)})
\end{equation}

Rectified Linear Unit (ReLU) is the profoundly utilised activation function within the convolutional blocks, as it is, in essence, a 'max-operation', making it computationally lightweight for its mathematical chemistry, as expressed via (4) in comparison to the sigmoid and TanH expressed via (5) and (6), respectively.

\begin{equation}
f(x) = \max(0, x)
\end{equation}

\begin{equation}
x \rightarrow \frac{1}{1 + e^{-x}}
\end{equation}

\begin{equation}
\tanh(x) = 2 \sigma(2x) - 1
\end{equation}

\begin{figure*}[h]
\centering
\includegraphics[width=1\columnwidth]{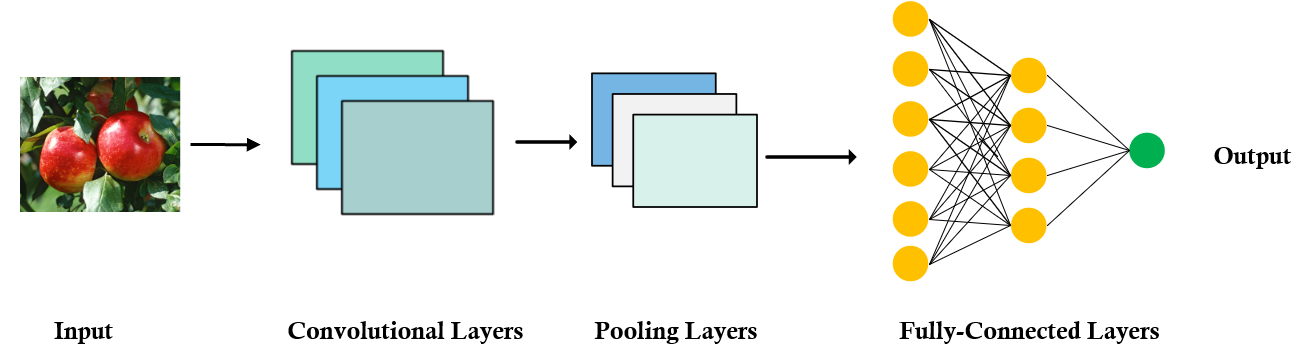}
\caption{The General structure of a CNN, highlighting convolutional layers, pooling, and fully connected layers.}
\label{Figure:1}
\end{figure*}

The visual manifestation of an abstract CNN is presented in Figure \ref{Figure:1}. The key components of CNNs can be labelled as a set of convolutional blocks consisting of filters that can be optimised, followed by a defined number of fully connected layers leading to the output~\cite{35}.

\section{Object Detection}

The development of effective object detectors presents several challenges for researchers and practitioners. A primary concern is handling variations in image resolutions and aspect ratios, which becomes more challenging when target objects exhibit significant differences in spatial dimensions. Class imbalance, especially in scenarios where obtaining an adequate number of images for certain classes is difficult, can negatively impact the performance of object detectors, resulting in biased predictions~\cite{36}.

\begin{figure}[h]
\centering
\includegraphics[width=1\columnwidth]{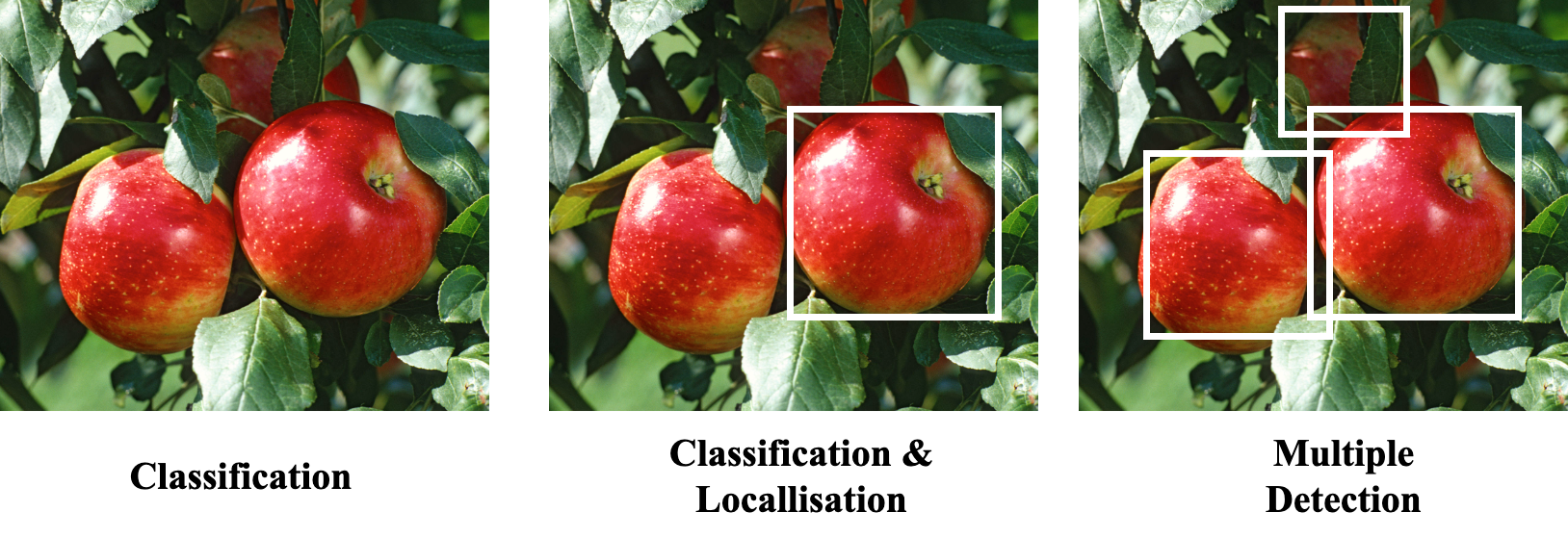}
\caption{Single and multiple objects in an image: Classification, Localization, Segmentation.}
\label{Figure:2}
\end{figure}

Another significant challenge lies in the computational complexity of object detection architectures, demanding substantial computational resources in terms of power, memory, and time~\cite{37,38}. Figure \ref{Figure:2} illustrates object detection for single and multiple objects in an image, depicting detectors with deep internal networks that require significant computational resources for processing complex image datasets and extracting crucial features.

\begin{figure*}[b]
\centering
\includegraphics[width=1.\columnwidth]{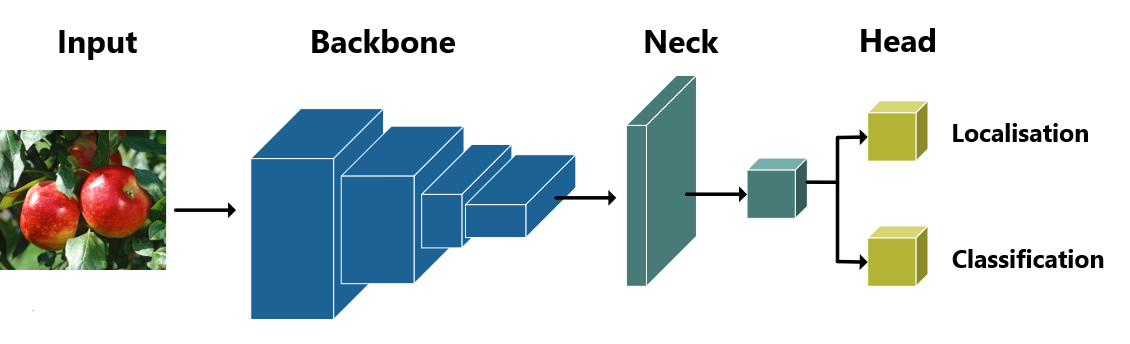}
\caption{Abstract architecture of single-stage object detectors.}
\label{Figure:3}
\end{figure*}

Object detection methods can be categorized into Two-stage object detectors and single-stage detectors. The former proposes candidate regions in the image, followed by classification and localization within those proposed regions. Examples of two-stage detectors include RCNN (Region-based Convolutional Neural Network)~\cite{xie2021oriented}, Fast R-CNN~\cite{girshick2015fast}, Faster R-CNN~\cite{ren2017faster}, and FPN (Feature Pyramid Network)~\cite{lin2017feature}.

RCNN~\cite{xie2021oriented}, introduced in 2014, utilized a selective search for candidate region proposals. It then employed a CNN network for feature extraction, followed by an SVM classifier for classification and localization. Although accurate, RCNN was computationally inefficient due to its two-stage process. Fast R-CNN~\cite{girshick2015fast} addressed efficiency concerns by introducing ROI pooling. This approach used ROI pooling to extract fixed-size feature maps for each region from the original feature maps, resulting in significant computational speed-up. Faster R-CNN~\cite{ren2017faster} improved upon Fast R-CNN by introducing the Region Proposal Network (RPN), which directly generated region proposals from convolutional feature maps, eliminating the need for a separate proposal stage. Integration of RPN into Fast R-CNN enhanced both speed and accuracy. FPN (Feature Pyramid Network)~\cite{lin2017feature} enhanced two-stage detectors by addressing the challenge of detecting targets at multiple scales. FPN generated a feature pyramid by incorporating feature maps of varying resolutions from different stages of the network, enabling the model to detect targets of different scales effectively.

While two-stage detectors exhibit impressive accuracy, their high computational demand limits their applications. Single-stage detectors aim to detect objects in a single pass, eliminating the need for a separate region proposal step, as shown in Figure \ref{Figure:3}. Notable single-stage detectors include SSD (Single Shot Multibox Detector), YOLO variants (You Only Look Once), RefineDet++, DSSD (Deconvolution Single Shot Detector), and RetinaNet. SSD~\cite{liu2016ssd} utilizes multiple convolutional feature maps at different scales for predicting bounding boxes and class probability scores. It effectively detects objects of various sizes and shapes in a single forward pass. RefineDet++\cite{sun2020dense} enhances the original RefineDet architecture by refining target proposals iteratively through multiple stages. Improved feature fusion mechanisms and refined target boundaries contribute to enhanced accuracy. DSSD (Deconvolution Single Shot Detector) incorporates deconvolution layers to retain spatial information lost during feature pooling. This aids in maintaining spatial resolution, allowing DSSD to capture fine-grained details. RetinaNet~\cite{39} addresses class imbalance with Focal Loss, assigning higher weights to hard, misclassified samples, improving the architecture's ability to handle class imbalance and enhance detection performance.

Single-stage detectors offer advantages such as faster inference speed and a lightweight footprint compared to two-stage detectors, making them suitable for resource-constrained environments. YOLO has emerged as a strong competitor among single-stage detectors, demonstrating impressive accuracy and real-time inference capabilities due to its straightforward architecture. It has proven to be effective in various real-world applications, showcasing its potential for production purposes.

\section{YOLO Architecture Background}

This section delves into the foundational principles and architecture underlying YOLO, detailing the unique advancements associated with each iteration. The YOLO algorithm, introduced in 2015 by Joesph Redmon et al.~\cite{17}, stands for "You Only Look Once." This name reflects its distinctive approach, examining the entire image just once to identify objects and their positions. In contrast to conventional methods employing two-stage detection processes, YOLO treats object detection as a regression problem~\cite{17}. In the YOLO paradigm, a single convolutional neural network is employed to predict bounding boxes and class probabilities for an entire image. This streamlined approach differs from traditional methods with more intricate pipelines.

\subsection{YOLOv1}
The core concept of YOLOv1 entails superimposing a grid cell-sized "s x s" onto an image. Whenever an object's centre lands within a grid cell, that specific cell is tasked with identifying the object, enabling other cells to ignore its existence in situations where multiple instances occur. Regarding object detection, every grid cell anticipates "B" bounding boxes, complete with dimensions and confidence scores. The confidence score indicates the probability of an object residing within the designated bounding box. Mathematically, the confidence score is represented as Equation (7):

\begin{equation}
\text{confidence score} = p(\text{object}) \times \text{IoU}_{\text{truth pred}}
\end{equation}

In this context, $p(\text{object})$ denotes the likelihood of the object's presence (with a range between 0 and 1), and IoU$_{\text{truth pred}}$ represents the intersection-over-union between the predicted bounding box and the ground truth.

The primary goal is accurately identifying and localising objects using bounding boxes. YOLO tackles challenges related to overlapping predicted bounding boxes through a non-maximum suppression (NMS) mechanism, eliminating those with an Intersection over Union (IoU) below a specified threshold. The original YOLO architecture, built on Darknet, introduced two sub-variants: one with 24 convolutional layers and another called 'Fast YOLO' with nine layers. Different penalties were assigned for bounding boxes containing objects and those indicating the absence of an object. The overall loss function incorporated coordinates, width, height, confidence score, and class probability considerations.

In terms of performance, the simpler YOLO version achieved a mean average precision (mAP) of 63.4\% at 45 frames per second (FPS), while the Fast YOLO variant reached 52.7\% at 155 FPS. Despite surpassing some real-time detectors, they fell short of state-of-the-art (SOTA) benchmarks at that particular time. Nevertheless, limitations such as lower recall and localization errors spurred further advancements in subsequent YOLO variants.

\subsection{YOLOv2}
Expanding on the achievements of YOLOv1, YOLOv2 brings forth notable enhancements in its design. This version incorporates ideas from Network-In-Network and VGG, opting for the Darknet-19 framework, consisting of 19 convolutional layers and 5 layers specifically designated for maximum pooling, as detailed in Table \ref{tab:darknet19}. YOLOv2 employs a blend of pooling layers and 1 x 1 convolutions, enabling down-sampling within the network architecture.

\begin{table}[h]
\centering
\caption{Darknet-19 framework}
\label{tab:darknet19}
\begin{tabular}{|c| c| c| c|}
\hline
Type & Filters & Size/Stride & Output \\
\hline
Convolutional & 32 & 3 x 3 & 224 x 224 \\
Maxpool & 2 x 2/2 & 112 x 112 & \\
Convolutional & 64 & 3 x 3 & 112 x 112 \\
Maxpool & 2 x 2/2 & 56 x 56 & \\
Convolutional & 128 & 3 x 3 & 56 x 56 \\
Convolutional & 64 & 1 x 1 & 56 x 56 \\
Convolutional & 128 & 3 x 3 & 56 x 56 \\
Maxpool & 2 x 2/2 & 28 x 28 & \\
Convolutional & 256 & 3 x 3 & 28 x 28 \\
Convolutional & 128 & 1 x 1 & 28 x 28 \\
Convolutional & 256 & 3 x 3 & 28 x 28 \\
Maxpool & 2 x 2/2 & 14 x 14 & \\
Convolutional & 512 & 3 x 3 & 14 x 14 \\
Convolutional & 256 & 1 x 1 & 14 x 14 \\
Convolutional & 512 & 3 x 3 & 14 x 14 \\
Convolutional & 256 & 1 x 1 & 14 x 14 \\
Convolutional & 512 & 3 x 3 & 14 x 14 \\
Maxpool & 2 x 2/2 & 7 x 7 & \\
Convolutional & 1024 & 3 x 3 & 7 x 7 \\
Convolutional & 512 & 1 x 1 & 7 x 7 \\
Convolutional & 1024 & 3 x 3 & 7 x 7 \\
Convolutional & 512 & 1 x 1 & 7 x 7 \\
Convolutional & 1024 & 3 x 3 & 7 x 7 \\
Convolutional & 1000 & 1 x 1 & 7 x 7 \\
Avgpool & Global & 1000 & \\
Softmax & & & \\
\hline
\end{tabular}
\end{table}

An essential challenge in object detection lies in the limited availability of labelled data, often confining methods to predetermined categories. YOLOv2 tackles this challenge by amalgamating the ImageNet and COCO datasets, broadening its detection capabilities to encompass over 9418 object instances~\cite{61}. For enhanced scalability, YOLOv2 adopts Word-Tree, a hierarchical classification and detection approach adept at efficiently handling the expanded array of categories.

Despite initial difficulties with small object detection, YOLOv2 marks substantial improvements over its predecessor. It introduces diverse data augmentation techniques and optimization strategies, yielding noteworthy advancements:

\begin{itemize}
  \item YOLOv2 predicts object dimensions across a range of sizes, from 320 x 320 to 608 x 608, by discarding fully connected layers present in YOLOv1.
  \item A 4\% increase in mean average precision (mAP) is attained through a higher resolution classifier. In contrast to V1, YOLOv2 undergoes training on 448 x 448 images for classification before detection, enhancing the accuracy of bounding box predictions.
  \item The integration of Batch Normalization addresses inconsistencies in input distribution during training, resulting in an approximate 2\% mAP improvement.
  \item Enhanced bounding box coordinate prediction is achieved by predicting location coordinates in relation to grid cell locations, leading to a 5\% increase in mAP with more uniform bounding box aspect ratios and sizes.
  \item YOLOv2 employs convolutional layers for feature extraction and predicts bounding boxes using anchor boxes, contributing to a 7\% improvement in recall.
  \item The adoption of a clustering algorithm based on K-means eliminates the need for manual selection of anchor boxes, thereby enhancing accuracy.
  \item To tackle the challenge of smaller object detection, skip connections inspired by ResNet are incorporated, resulting in a 1\% increase in mAP. For instance, a 26 x 26 x 512 feature map transforms into a 13 x 13 x 2048 feature map, concatenating with the model's output, enabling more robust object recognition across various dimensions.
\end{itemize}

\subsection{YOLOv3}
The introduction of YOLOv3~\cite{Borisyuk2018} in 2018 by Joesph Redmon et al. represented a significant evolution, featuring an expanded architecture outlined in Table \ref{tab:yolov3}. This iteration embraced contemporary technological advancements while maintaining real-time processing capabilities. Similar to YOLOv2, YOLOv3 predicts four coordinates for each bounding box but introduces an objectness score for each box, determined through logistic regression. This score assumes values of 1 or 0, indicating whether the anchor box has the highest overlap with the ground truth (1) or other anchor boxes (0). Unlike Faster R-CNN~\cite{Won2019}, YOLOv3 associates a single anchor box with each ground truth object, incurring only the classification loss in cases where no anchor box is associated, excluding localization and confidence losses.

\begin{figure}[h]
  \centering
  \includegraphics[width=0.7\columnwidth]{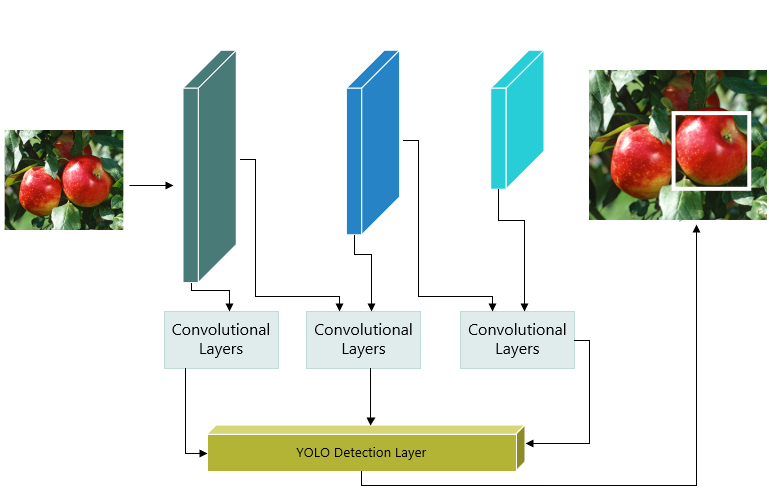}
  \caption{Multi-scale Detection Architecture}
  \label{Figure:4}
\end{figure}

\begin{table}[h]
  \centering
  \caption{YOLOv3 Architecture}
  \label{tab:yolov3}
  \begin{tabular}{|c|c|c|c|c|}
    \hline
    \textbf{Layer} & \textbf{Filters} & \textbf{Size} & \textbf{Repeat} & \textbf{Output Size} \\ \hline
    Image & --- & --- & --- & 416 $\times$ 416 \\
    Conv & 32 & 3 $\times$ 3 / 1 & 1 & 416 $\times$ 416 \\
    Conv & 64 & 3 $\times$ 3 / 2 & 1 & 208 $\times$ 208 \\
    Conv & 32 & 1 $\times$ 1 / 1 & Conv $\times$ 1 & 208 $\times$ 208 \\
    Conv & 64 & 3 $\times$ 3 / 1 & Conv $\times$ 1 & 208 $\times$ 208 \\
    Residual & --- & --- & Residual $\times$ 1 & 208 $\times$ 208 \\
    Conv & 128 & 3 $\times$ 3 / 2 & 1 & 104 $\times$ 104 \\
    Conv & 64 & 1 $\times$ 1 / 1 & Conv $\times$ 2 & 104 $\times$ 104 \\
    Conv & 128 & 3 $\times$ 3 / 1 & Conv $\times$ 2 & 104 $\times$ 104 \\
    Residual & --- & --- & Residual $\times$ 2 & 104 $\times$ 104 \\
    Conv & 256 & 3 $\times$ 3 / 2 & 1 & 52 $\times$ 52 \\
    Conv & 128 & 1 $\times$ 1 / 1 & Conv $\times$ 8 & 52 $\times$ 52 \\
    Conv & 256 & 3 $\times$ 3 / 1 & Conv $\times$ 8 & 52 $\times$ 52 \\
    Residual & --- & --- & Residual $\times$ 8 & 52 $\times$ 52 \\
    Conv & 512 & 3 $\times$ 3 / 2 & 1 & 26 $\times$ 26 \\
    Conv & 256 & 1 $\times$ 1 / 1 & Conv $\times$ 8 & 26 $\times$ 26 \\
    Conv & 512 & 3 $\times$ 3 / 1 & Conv $\times$ 8 & 26 $\times$ 26 \\
    Residual & --- & --- & Residual $\times$ 8 & 26 $\times$ 26 \\
    Conv & 1024 & 3 $\times$ 3 / 2 & 1 & 13 $\times$ 13 \\
    Conv & 512 & 1 $\times$ 1 / 1 & Conv $\times$ 4 & 13 $\times$ 13 \\
    Conv & 1024 & 3 $\times$ 3 / 1 & Conv $\times$ 4 & 13 $\times$ 13 \\
    Residual & --- & --- & Residual $\times$ 4 & 13 $\times$ 13 \\ \hline
  \end{tabular}
\end{table}

In contrast to utilizing SoftMax for classification, YOLOv3 employs binary cross-entropy, enabling the assignment of multiple labels to a single box. The architecture integrates an extensive feature extractor with 53 convolutional layers and incorporates residual connections.

Significant improvements involve a modified spatial pyramid pooling (SPP) block within the backbone to accommodate a broader receptive field. YOLOv3 organizes feature maps into three scales: (416$\times$416), (13$\times$13), (26$\times$26), and (52$\times$52) for input, each featuring three prior boxes for every position (as depicted in Figure \ref{Figure:4}). Collectively, these enhancements resulted in a 2.7\% improvement in the AP50 metric.

The determination of eight prior boxes, distributed across the three-scale feature maps, employs the K-means algorithm. Larger-scale feature maps incorporate smaller precedent boxes. The foundational architecture of YOLOv3 referred to as Darknet-53, replaces all max-pooling layers with stride convolutions and integrates residual connections. Comprising 53 convolutional layers (as detailed in Table \ref{tab:yolov3}), this backbone architecture emerged as the primary benchmark for object detection shifted from PASCAL VOC~\cite{Everingham2010} to Microsoft COCO ~\cite{Lin2014}. Consequently, all subsequent YOLO models were evaluated using the MS COCO dataset. YOLOv3 achieved notable results: an average precision (AP) of 36.2\% and an AP-50 of 60.6\% at a processing speed of 20 frames per second (FPS), surpassing the pace of previous state-of-the-art models.

\subsection{YOLOv4}

\begin{figure}[h]
\centering
\includegraphics[width=1\columnwidth]{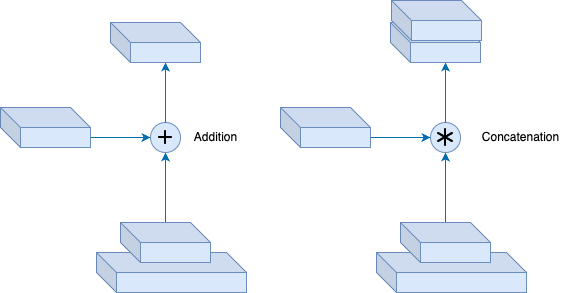}
\caption{Path Aggregation YOLOv4 (a) Addition (b) Concatenation}
\label{Figure:5}
\end{figure}

In April 2020, a team led by Alexey Bochkovskiy introduced YOLOv4~\cite{63}, representing a profound departure from its predecessors with revolutionary architectural changes aimed at improving performance while maintaining real-time capabilities. The key advancements in YOLOv4 include the integration of CSP Darknet53, SPP structure~\cite{He}, PANet architecture~\cite{Ma2019} (depicted in Figure \ref{Figure:5}), CBN integration~\cite{Yao2021}, and SAM incorporation~\cite{He2023}, resulting in an efficient and robust object detection model. Designed to simplify the training of object detectors, YOLOv4 aims to be accessible to individuals with varying technical expertise. The study also validated the effectiveness of state-of-the-art methodologies, such as bag-of-freebies and bag-of-specials, to enhance the efficiency of the training pipeline.

Unlike YOLOv3, where a single anchor point detected a ground truth, YOLOv4 uses multiple anchor points for a single ground truth detection. This approach improves the selection ratio of positive samples, reduces the imbalance between positive and negative samples, and enhances boundary detection accuracy.

YOLOv4 employs the Complete Intersection over Union (CIoU) loss, depicted in Equation (8), to refine localization accuracy by incorporating factors like IoU, maximum IoU, and regularization. The utilization of this loss function enhances YOLOv4's capability to accurately locate and outline objects in images, thereby improving overall object detection performance.

\begin{equation}
\begin{aligned}
& L_{CIoU} = 1 - \text{IoU}(b, \hat{b}) + \frac{(\rho^2 - \text{IoU}(b, \hat{b})^2)}{\rho^2} \\
& \quad + \alpha \cdot \frac{v}{(1 - \text{IoU}(b, \hat{b}) + v)}
\end{aligned}
\end{equation}

Here, $L_{CIoU}$ denotes the CIoU loss, $b$ stands for the predicted bounding box, $\hat{b}$ represents the ground truth bounding box, $\text{IoU}(b, \hat{b})$ computes the Intersection over Union (IoU) between the predicted and ground truth boxes, $\rho^2$ serves as a parameter for the maximum possible IoU, $\alpha$ acts as a balancing factor, and $v$ is employed to address small bounding boxes.

\subsection{YOLOv5} 

In 2020, Glenn Jocher introduced YOLOv5, following the release of YOLOv4~\cite{Yao2021}. YOLOv5, managed by Ultralytics, takes a different path from YOLOv4 in several key aspects. Notably, YOLOv5 opts for PyTorch instead of Darknet for development, expanding its user base due to PyTorch's user-friendly characteristics. YOLOv5 incorporates various enhancements to improve its performance in object detection. At its core, YOLOv5 features a CSP (Cross Stage Partial) Net, derived from the ResNet architecture, which includes a cross-stage partial connection for enhanced network efficiency. The CSPNet is complemented by multiple SPP (Spatial Pyramid Pooling) blocks for feature extraction at different scales.

The architecture's neck includes a PAN (Path Aggregation Network) module and subsequent upsampling layers to improve feature map resolution~\cite{Solawetz2020}. The head of YOLOv5 utilizes convolutional layers to predict bounding boxes and class labels. YOLOv5 employs anchor-based predictions, associating each bounding box with predetermined anchor boxes of specific shapes and sizes. The loss function in YOLOv5 combines Binary Cross-Entropy and Complete Intersection over Union (CIoU) for class, objectness, and localization losses, expressed as (9):

\begin{equation}
\text{loss} = \lambda_1 \cdot L_{\text{cls}} + \lambda_2 \cdot L_{\text{obj}} + \lambda_3 \cdot L_{\text{loc}}
\end{equation}

Where $L_{\text{cls}}$, $L_{\text{obj}}$, and $L_{\text{loc}}$ denote the Binary Cross-Entropy loss for class predictions, Binary Cross-Entropy loss for objectness predictions, and CIoU loss for localization, respectively. The $\lambda$ values serve as weighting factors for each loss component.

The primary objective of YOLOv5 is to enhance efficiency and accuracy, surpassing its predecessors. It brings advancements in feature extraction, feature aggregation, and anchor-based predictions. Furthermore, it ensures a seamless transition from PyTorch to ONNX and CoreML frameworks, enhancing compatibility with iOS devices. In evaluations on the MS COCO dataset's test-dev 2017 split, YOLOv5x achieved an average precision (AP) score of 50.7\% with a 640-pixel image size, processing at a high speed of 200 frames per second (FPS) on an NVIDIA V100. With a larger input size of 1536 pixels, YOLOv5 achieved an even higher AP score of 55.8\%, as evident from Table 3.

\begin{table}[h]
\centering
\caption{Variant Comparison of YOLOv5}
\label{tab:yolov5}
\begin{tabularx}{\textwidth}{XXXX}
\hline
Model & Average Precision (@50) & Parameters & FLOPs \\ \hline
YOLO-v5s & 55.8\% & 7.5M & 13.2B \\
YOLO-v5m & 62.4\% & 21.8M & 39.4B \\
YOLO-v5l & 65.4\% & 47.8M & 88.1B \\
YOLO-v5x & 66.9\% & 86.7M & 205.7B \\ \hline
\end{tabularx}
\end{table}

\subsection{YOLOv6} 

Released in September 2022 by the Meituan Vision AI Department, YOLOv6 is a single-stage object detection framework designed specifically for industrial applications. This version brings significant improvements and architectural refinements, notably introducing CSPDarknet as the new backbone architecture, surpassing its predecessors' efficiency and speed benchmarks, YOLO-v4 and YOLO-v5. One key enhancement in YOLO-v6 is the integration of a feature pyramid network (FPN), expanding the range of feature scales and resulting in a noticeable improvement in detection accuracy. This underscores the commitment to enhancing overall performance~\cite{Wang2022}. 

\begin{figure*}[h]
\centering
\includegraphics[width=1\columnwidth]{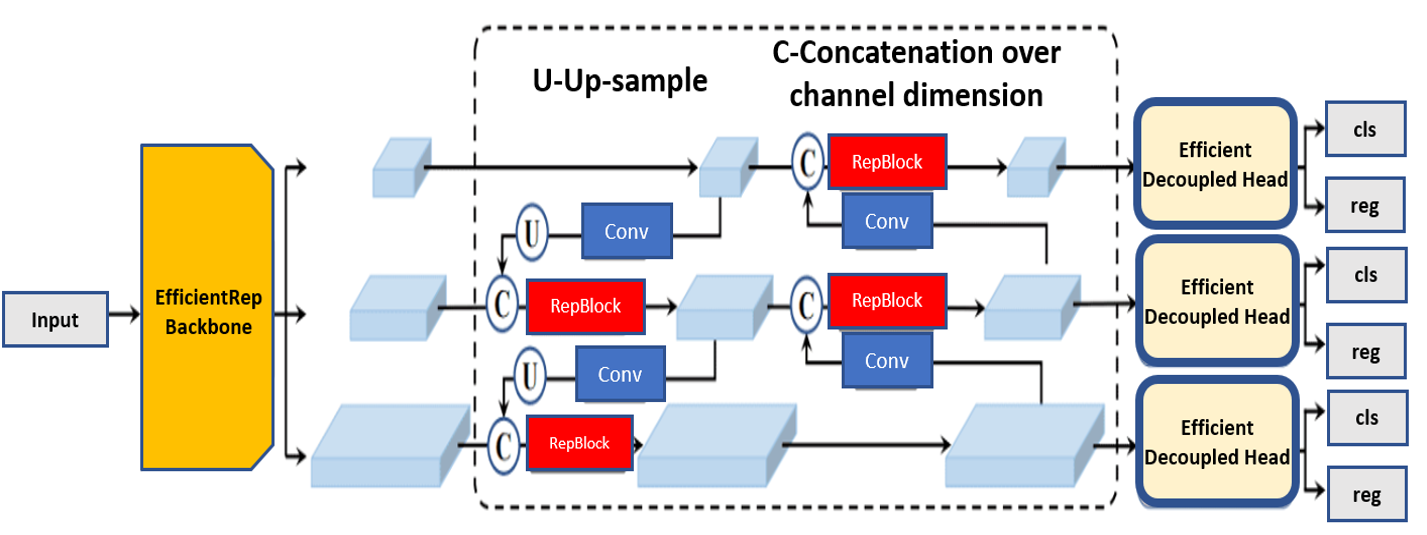}
\caption{PANet Configuration~\cite{hussain2023yolo}}
\label{Figure:6}
\end{figure*}

YOLO-v6 is meticulously crafted for optimal real-time object detection performance, demonstrating impressive frame rates on both central processing units (CPUs) and graphics processing units (GPUs).A pivotal evolution in the YOLOv6 architecture involves the separation of the classification and box regression heads, as illustrated in Figure \ref{Figure:6}. This strategic architectural revision introduces additional layers within the network, effectively segregating these crucial functions from the final head~\cite{Wang2023}. Empirical evidence supports the impact of this refinement in elevating the overall model's performance, reinforcing its capabilities~\cite{Solawetz2022}.

In aggregate, YOLOv6 marks a substantial advancement in the progression of YOLO architectures, incorporating a wide range of enhancements that span speed, accuracy, and operational efficiency. A thorough assessment of the MS COCO dataset's test-dev 2017 subset highlighted the capabilities of the YOLOv6L model, delivering an average precision (AP) of 52.5\% and an AP50 of 70\%. Notably, this commendable performance was attained while sustaining a processing speed of around 50 frames per second (FPS) on an NVIDIA Tesla T4 GPU. YOLOv6 is presented in three distinct variants, as outlined in Table 4. Notably, YOLOv6nano stands out as the smallest and fastest alternative, characterized by a minimal parameter count. This feature makes it particularly well-suited for real-time object detection tasks on devices with limited computational capabilities. Advancing to YOLO-v6tiny, this variant offers a more expansive architecture compared to YOLOv6nano, resulting in increased accuracy, as evidenced in Table 4. YOLOv6tiny proves valuable when precision is crucial, especially in scenarios involving the detection of smaller objects.

\begin{table}[h]
\centering
\caption{YOLOv6 Variant Comparison}
\label{tab:yolov6}
\begin{tabularx}{\textwidth}{XXXXX}
\hline
v6 Variant & Size & mAP & Parameters & FLOPs \\
\hline
Nano & 416-640 & 30.8-35.0\% & 4.3M & 4.7-11.1G \\
Tiny & 640 & 41.3\% & 15M & 36.7G \\
Small & 640 & 43.1\% & 17.2M & 44.2G \\
\hline
\end{tabularx}
\end{table}
Conversely, YOLOv6small leads in architectural complexity, delivering a higher level of accuracy. This configuration is particularly suitable for scenarios where detecting smaller objects within the visual field is paramount. The selection among these variants depends on the specific use case and available computational resources. YOLOv6nano is an optimal choice for scenarios requiring real-time detection on low-powered devices, while preferences for YOLOv6tiny or YOLOv6small may arise in instances where greater accuracy and the identification of smaller objects are essential. The decision should be tailored to the available resources and the desired accuracy threshold.

\subsection{YOLOv7} 

Introduced in July 2022, YOLOv7~\cite{Xu2022} marked a significant leap forward from its predecessors, showcasing improved accuracy and speed enhancements ranging from 5 FPS to 160 FPS. These advancements primarily centred on boosting efficiency and scalability by integrating the Extended Efficient Layer Aggregation Network (E-ELAN)~\cite{Ding2021} and implementing a scalable concatenation-based architecture. E-ELAN plays a pivotal role in managing the gradient path, thereby augmenting model learning and convergence. This technique is versatile and applicable to models with stacked computational blocks, adeptly shuffling and merging features from different groups while preserving the integrity of the gradient path. Model scaling is another critical aspect of YOLOv7, enabling the creation of models of varying sizes. The devised scaling strategy adjusts the depth and width of the blocks uniformly, maintaining the optimal model structure while mitigating hardware resource consumption. The amalgamation of various techniques, collectively termed "bag-of-freebies," further amplifies YOLOv7's performance. One such technique mirrors the re-parameterized convolution concept employed in YOLOv6. However, the RepConvN approach was introduced in YOLOv7 due to identified issues with the identity connection in RepConv~\cite{Huang2016} and concatenation in DenseNet~\cite{Jocher2023}.

Additionally, YOLOv7 utilizes coarse label assignment for the auxiliary head, reserving fine label assignment for the lead head. While the auxiliary head contributes to the training process, the lead head produces the final output, as shown in Figure \ref{Figure:6}. Moreover, batch normalization is employed, amalgamating the mean and variance of batch normalization into the convolutional layer's bias and weight during inference, ultimately enhancing model performance~\cite{Solawetz2022b}. Under rigorous evaluation on the MS COCO dataset's test-dev 2017, YOLOv7E6 demonstrated outstanding performance, achieving an average precision (AP) of 55.9\% and an AP for the IoU threshold of 0.5 (AP50) of 73.5\%, as showcased in Table 5.

\begin{table}[h]
\centering
\caption{YOLOv7 Variant Comparison}
\label{tab:yolov7}
\begin{tabularx}{\textwidth}{XXXXX}
\hline
v7 Variant & Size & mAP & Parameters & FLOPs \\
\hline
YOLO-v7 tiny & 640 & 52.8\% & 6.2M & 5.8G \\
YOLO-v7 & 640 & 69.7\% & 36.9M & 104.7G \\
YOLO-v7X & 640 & 71.1\% & 71.3M & 189.9G \\
YOLO-v7E6 & 1280 & 73.5\% & 97.2M & 515.2G \\
YOLO-v7D6 & 1280 & 73.8\% & 154.7M & 806.8G \\
\hline
\end{tabularx}
\end{table}

\subsection{YOLOv8} 
In January 2023, Ultralytics introduced YOLOv8, making a significant entrance into the field of computer vision~\cite{Solawetz2023}. The model's precision was extensively assessed through evaluations on both COCO and Roboflow 100 datasets~\cite{Solawetz2023}. YOLOv8 distinguishes itself with user-oriented features, including a user-friendly command-line interface and a well-organized Python package. The supportive YOLO community further enhances the model's accessibility for users. The innovation in YOLOv8, as detailed in its methodology~\cite{Jocher2020}, diverges from traditional anchor-based methods. Instead of relying on predetermined anchor boxes, YOLOv8 adopts an anchor-free approach by predicting the object's centre. This adjustment addresses challenges associated with anchor boxes that may not accurately represent custom dataset distributions. This approach's advantages include reducing the number of box predictions and expedited post-processing steps involving Non-Maximum Suppression. Notably, YOLO-v8's training routine, which incorporates techniques like online image augmentation, including mosaic augmentation, enhances the model's ability to detect objects across diverse conditions and novel spatial arrangements.

In its architectural evolution from its predecessor, YOLOv5 (also authored by the same individuals), YOLOv8 introduces changes across its components. For instance, in the neck segment, YOLOv8 directly concatenates features without enforcing uniform channel dimensions. This strategy contributes to a reduction in parameter count and overall tensor size. When assessed on the MS COCO dataset's test-dev 2017 subset, YOLOv8x demonstrated an average precision (AP) of 53.9\% at an image size of 640 pixels, surpassing YOLOv5's AP of 50.7\% with the same input size. Furthermore, YOLOv8x exhibited remarkable processing speed, achieving 280 frames per second (FPS) using an NVIDIA A100 with TensorRT. Notably, YOLOv8 is available in five distinct variants, each tailored to specific accuracy and computational requirements, as detailed in Table 6.

\begin{table}[h]
\centering
\caption{YOLO-v8 Variant Comparison}
\label{tab:yolov8}
\begin{tabularx}{\textwidth}{XXXXX}
\hline
Model & Size & mAP & Parameters & FLOPs \\
\hline
YOLO-v8n & 640 & 37.3\% & 3.2M & 8.7G \\
YOLO-v8s & 640 & 44.9\% & 11.2M & 28.6G \\
YOLO-v8m & 640 & 50.2\% & 25.9M & 78.9G \\
YOLO-v8l & 640 & 52.9\% & 43.7M & 165.2G \\
YOLO-v8x & 640 & 53.9\% & 68.2M & 257.8G \\
\hline
\end{tabularx}
\end{table}

\subsection{YOLOv9}
YOLOv9, released in February 2024 ~\cite{wang2024yolov9} represents the latest addition to the mainstream YOLO variants. YOLOv9 boasts two key innovations: the Programmable Gradient Information (PGI) framework and the Generalized Efficient Layer Aggregation Network (GELAN). The PGI framework aims to address the issue of information bottlenecks, inherent in deep neural networks in addition to enabling deep supervision mechanisms to be compatible with lightweight architectures. By implementing PGI, both lightweight and deep architectures can leverage substantial improvements in accuracy, as PGI mandates reliable gradient information during training, thus enhancing the architecture's capacity to learn and make accurate predictions.

The GELAN architecture was purposefully designed to boost performance in object detection tasks via high efficiency and lightweight footprint. GELAN manifests high performance across varying computational blocks and depth configurations, making it suitable for deployment on different inference devices, including resource-constrained edge devices. By combining the above two frameworks (PGI and GELAN), YOLOv9 presents a significant advancement in lightweight object detection. Although in its early days, YOLOv9 has achieved remarkable competitiveness in object detection tasks, outperforming YOLOv8 concerning parameter reduction and computational efficiency while improving Average Precision (AP) by 0.6\% on the MS COCO dataset. 

\subsection{YOLOv10}
Building upon the momentum of innovation in the YOLO series, another groundbreaking development emerged in the same year with the release of YOLOv10, also in 2024. This version pushes the boundaries further by addressing the challenges associated with real-time object detection, which is critical for applications requiring rapid and accurate responses, such as agricultural monitoring and autonomous vehicle navigation.

YOLOv10 distinguishes itself by completely eliminating the reliance on non-maximum suppression (NMS) during post-processing, which is a significant step forward in enhancing inference speed. This model adopts a novel NMS-free training approach using dual label assignments, allowing for a harmonious integration of accuracy and speed by ensuring that the model remains computationally efficient while still capturing essential detection features. Moreover, the architectural enhancements in YOLOv10 include the implementation of lightweight classification heads, spatial-channel decoupled downsampling, and rank-guided block design, each contributing to substantial reductions in computational demands and parameter count. These innovations not only improve the model's efficiency but also its scalability across various devices, from high-power servers to resource-limited edge devices.

Extensive testing demonstrates that YOLOv10 sets a new benchmark for the performance-efficiency trade-off. It achieves remarkable improvements in latency and model size reduction compared to YOLOv9, while still delivering competitive or superior detection accuracy. This is particularly evident in its application to the COCO dataset, where YOLOv10 shows notable advancements in detection metrics, solidifying its position as a leader in the field of real-time object detection technologies \cite{wang2024yolov10}. 

Table 7 provides a comparative overview of the major YOLO variants up to the current date. The table illustrates the iterative evolution of the YOLO series of object detectors, with each iteration advancing the state-of-the-art in computer vision.

\begin{table}[h]
\centering
\caption{YOLO Variant Comparison}
\label{tab:yolo-variants}
\begin{tabularx}{\textwidth}{ccXc}
    \toprule
    \textbf{Version} & \textbf{Date} & \textbf{Contributions} & \textbf{Framework} \\ \midrule
    v1 & 2015  & One-shot object detector & Darknet \\ \midrule
    v2 & 2016  & Multi-scale training, dimensional clustering & Darknet \\ \midrule
    v3 & 2018  & SPP block, Darknet-53 & Darknet \\ \midrule
    v4 & 2020  & Mish-based activation, CSPDarknet-53 backbone & Darknet \\ \midrule
    v5 & 2020  & Anchor-free detection, SWISH-based activation, PANet & PyTorch \\ \midrule
    v6 & 2022  & Self-attention, anchor-free object detection & PyTorch \\ \midrule
    v7 & 2022  & Transformers, E-ELAN reparameterization & PyTorch \\ \midrule
    v8 & 2023  & GANs, anchor-free detections & PyTorch \\ \midrule
    v9 & 2024  & Programmable Gradient Information (PGI), Generalized Efficient Layer Aggregation Network (GELAN) & PyTorch \\ \hline
    v10 & 2024  & NMS-free training approach, dual label assignments, holistic model design for enhanced accuracy and efficiency & PyTorch \\ \bottomrule
\end{tabularx}
\end{table}

\section{Agricultural Applications of YOLO}

In this section, we provide a comprehensive review of the current body of literature regarding the utilization of different YOLO variants for various agricultural applications. The discussion is tailored to include applications such as weed detection, crop classification, disease detection, animal tracking, and precision farming.

\subsection{Weed Detection Using YOLO}
Weed detection is a crucial aspect of modern agriculture as it directly impacts crop yield and resource optimization. Traditional methods of weed management can be labour-intensive and time-consuming. The integration of YOLO variants in weed detection brings forth a promising solution by offering real-time and efficient identification of weeds in agricultural landscapes.

This subsection explores the application of YOLO in weed detection, focusing on its potential to revolutionize weed management practices. We delve into the challenges associated with weed detection in traditional agriculture and examine how YOLO variants address these challenges. Additionally, we explore real-world scenarios where YOLO has demonstrated efficacy in accurately identifying and localizing weeds, contributing to advancements in precision farming.

\cite{RN1} developed and implemented a real-time weed detection system targeting green onion crops. Leveraging the YOLOv3 deep learning algorithm, the system, coined YOLO-WEED, demonstrated notable efficiency and precision in identifying weeds within video frames captured by Unmanned Aerial Vehicles (UAVs) \cite{RN1}. This positions YOLO-WEED as a valuable asset for precise agricultural activities such as targeted spraying and weed management. The system's performance evaluation, based on mean average precision and F1 score metrics, yielded impressive results with a mean average precision of 93.81\% and an F1 score of 0.94.
The notable swiftness and accuracy of the YOLO-WEED system make it a promising candidate for seamless integration into real-time automated aerial spray systems specifically designed for green onion fields \cite{RN1}. However, it is important to acknowledge certain limitations. The system's effectiveness is contingent upon the resolution of UAV video frames. Notably, the YOLOv3 algorithm employed in the system encounters challenges in detecting smaller objects, presenting difficulties in identifying diminutive weeds within the green onion fields \cite{RN1}.
Moreover, the YOLO-WEED system necessitates the presence of an onboard computer, introducing an additional weight factor to the UAV sprayer system. Despite these considerations, the system's overall performance underscores its potential as an impactful solution for advancing precision farming practices, particularly in the context of weed detection and control in green onion cultivation.

Boyu Ying et al. conducted a meticulous study on detecting weeds in images of carrot fields using an improved YOLOv4 model \cite{RN2}. The researchers collected test images from the carrot fields in Central China's Henan Province and aimed to detect four common field weeds: crabgrass, plantain, pale persicaria, and cephalanoplos. They developed a lightweight weed detection model called YOLOv4-weeds, which replaced the backbone network of YOLOv4 with MobileNetV3-Small. This modification reduced the memory requirement of image processing and improved the efficiency and accuracy of small weed detection in complex environments. The authors conducted comparative experiments with other detection models and demonstrated that the YOLOv4-weeds model outperformed these models, especially in detecting diverse weeds in complex field scenes.
The research outcomes provide valuable insights and a reference for weed detection, robot weeding, and selective spraying in agricultural settings. However, the study has potential limitations, such as the diversity of weed species, generalization to other crops, robustness to environmental variability, real-world deployment and validation, and consideration of computational resources. These findings significantly affect the agricultural sector, as weed management is critical to crop yield and quality. The study provides a framework for improving weed detection in complex agricultural environments, which can lead to efficient and cost-effective weed management practices. However, further research is required to address the study's limitations and enhance the generalizability of the findings to other crops and environmental settings.

\begin{figure*}[h]
\centering
\includegraphics[width=1\columnwidth]{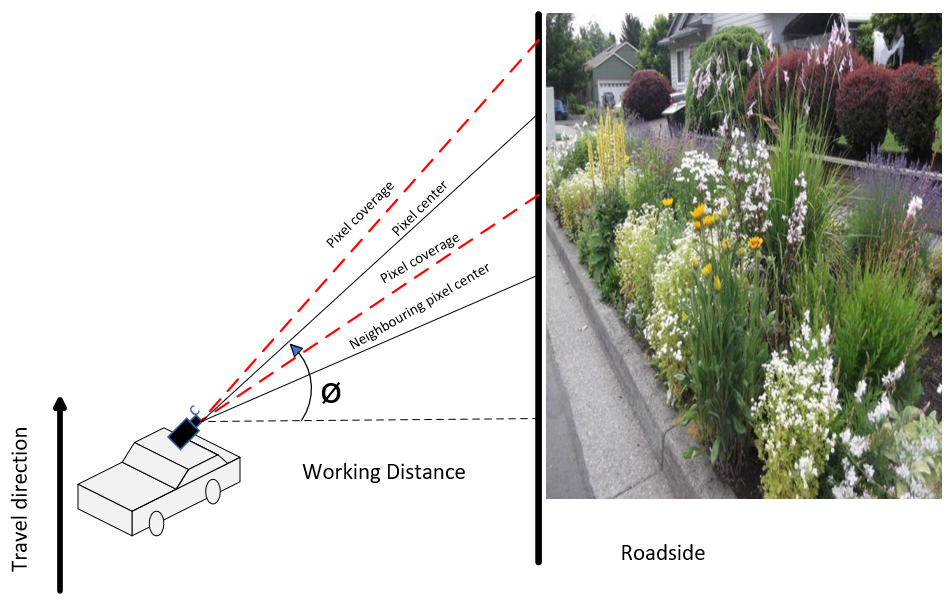}
\caption{Plant Species Detection On Device Deployment}
\label{Figure:7}
\end{figure*}

Dyrmann et al. conducted a study focusing on the identification and tracking of invasive alien plant species (IAPS) along state roads, utilizing a camera-based monitoring system \cite{RN3}, as shown in Figure \ref{Figure:7}. The employed deep learning algorithms successfully detected and classified IAPS in the collected images, demonstrating promising results for real-time mapping of invasive alien plant species at driving speeds of 110 km/h. Despite notable achievements, the study acknowledges certain limitations, such as challenges in detecting specific plant species like Lupinus polyphyllus and Pastinaca sativa, and emphasizes the need for addressing these issues in future research. The study provides valuable insights for cost-effective and efficient environmental conservation along roadsides.

Chen et al. concentrated on developing a YOLO-sesame model for weed detection in sesame fields \cite{RN4}. The model, incorporating an improved attention mechanism and feature fusion, exhibited high performance in terms of Frames Per Second (FPS) and mean Average Precision (mAP). While showcasing promising results, the study highlights regional specificity in the dataset and emphasizes the necessity for further work to enhance the model's applicability to embedded devices. This work contributes to the advancement of weed detection methodologies in sesame cultivation.

Wang et al. conducted field trials in Zhangjiakou City, Hebei Province, to test their method for real-world application \cite{RN5}. The study's technique for image preprocessing and the deployment of an improved YOLOv5 CNN showed satisfactory performance in practical scenarios. Despite notable achievements, the study recognizes limitations related to environmental conditions and underscores the importance of addressing these challenges for broader applicability. The study contributes to the progress of weed detection methodologies with potential implications for the early-stage identification of invasive weeds.

Costello et al. investigated the use of RGB and HSI data for the field mapping of Parthenium weed in a controlled environment \cite{RN6}. Employing deep-learning algorithms, including decision-tree-based protocols, the study achieved high accuracy in detecting and categorizing Parthenium weed growth stages. However, the study acknowledges limitations such as the controlled environment and emphasizes the need for further exploration of AI algorithms and technology improvements for enhanced detection success. This research offers valuable insights into the application of AI and imaging techniques for Parthenium weed detection, with potential implications for weed management and agriculture.

\begin{table*}[!b]
\caption{Studies on Weed Detection Using YOLO}
\label{tab:weed_detection}
\centering
\resizebox{\textwidth}{!}{%
\begin{tabular}{p{1.5cm}p{4.5cm}p{6.5cm}}
\hline
\textbf{Authors} & \textbf{Model} & \textbf{Details} \\
\hline
\cite{RN1} & YOLOv3 & Real-time weed detection with UAVs in green onion fields. Mean average precision of 93.81\%, F1 score of 0.94. \\
\cite{RN2} & YOLOv4 & Detection of weeds in carrot fields. Efficient detection of diverse weeds, outperforming other models in complex field scenes. \\
\cite{RN3} & YOLO & Identification and tracking of invasive alien plant species (IAPS) along Danish state roads. Successful results at driving speeds of 110 km/h. \\
\cite{RN4} & YOLO-custom & Weed detection in sesame fields with improved attention mechanism and feature fusion. High FPS and mAP. \\
\cite{RN5} & YOLOv5 & Field trials in Zhangjiakou City, Hebei Province, for weed detection. Satisfactory performance in practical scenarios. \\
\cite{RN6} & YOLOv4 & Field mapping of Parthenium weed in a controlled environment. High accuracy in detecting and categorizing Parthenium weed growth stages. \\
\cite{RN7} & YOLOv4 \& YOLOv5 & Introduction of YOLOWeeds benchmark dataset for detecting various weed types in cotton production systems. Evaluation of six different YOLO object detection models. \\
\cite{RN8} & YOLO & Detection of poppy (Papaver rhoeas) in wheat fields. Approximate 75\% accuracy. \\
\cite{RN9} & YOLO & Evaluation for weed detection in various turfgrass scenarios. Emphasis on challenges associated with accurate weed detection in turfgrass. \\
\cite{RN10} & YOLO-v3 & Novel method for weed detection in vegetable fields using a two-stage approach based on Convolutional Neural Networks (CNN). \\
\hline
\end{tabular}%
}
\end{table*}
Dang et al. introduced the YOLOWeeds benchmark dataset for detecting various weed types in cotton production systems \cite{RN7}. Evaluating six different YOLO object detection models, the study provides a detailed experimental setup, emphasizing challenges in weed management within cotton production. The results showcase the potential of YOLOv4 and YOLOv5 for real-time weed detection, prompting further exploration in machine vision-based weeding systems. This work contributes to the advancement of automated weed identification with potential applications in sustainable weed management in agriculture.

Pérez-Porras et al. conducted a study on detecting poppy (Papaver rhoeas) in wheat fields using YOLO architectures \cite{RN8}. Evaluating different YOLO models, the study optimized hyperparameters and assessed computational efficiency. Despite achieving an accuracy of approximately 75\%, the study acknowledges the need for field validation and integration with agricultural practices. This research provides valuable insights into early weed detection in agricultural fields, specifically in the case of poppy detection in wheat fields.

Sportelli et al. evaluated the performance of YOLO object detectors for weed detection in various turfgrass scenarios \cite{RN9}. Utilizing three datasets with specific characteristics, the study underscores the challenges associated with accurate weed detection in turfgrass. While achieving high performance, the study acknowledges limitations in model performance and emphasizes the need for further research to address these challenges. This work contributes to a comprehensive understanding of the trade-offs between model performance and computational efficiency in weed detection.

Jin et al. proposed a novel method using YOLO-v3 to identify weeds in vegetable fields \cite{RN10}. Adopting a two-stage approach based on Convolutional Neural Networks (CNN), the method accurately detects weeds by focusing on the detection of vegetable crops. Despite promising results, the study highlights the primary limitation in the lack of robust sensing technology for the commercial development of intelligent robotic weeders. This research provides a feasible approach for weed detection in various crops and emphasizes the need for advancements in sensing technology for broader applications.
Table \ref{tab:weed_detection} presents a compilation of studies employing diverse YOLO architectures for the detection of weeds in the agricultural sector.

\subsection{Crop Detection via YOLO}
In the domain of precision agriculture, crop detection plays a pivotal role in optimizing farming practices and resource utilization. Accurate identification and delineation of crops in large-scale fields contribute to improved monitoring, yield estimation, and resource management. Various studies have explored the application of YOLO variants to address the challenges associated with crop detection, showcasing the potential for real-time, efficient, and precise identification of different crop types.

This subsection provides an overview of notable research endeavours that employ YOLO models for crop detection. The studies discussed herein present advancements, methodologies, and outcomes related to the utilization of YOLO-based approaches in identifying and delineating crops in diverse agricultural settings. The significance of robust crop detection methodologies in the context of precision farming is emphasized, highlighting how YOLO variants contribute to addressing the evolving needs of modern agriculture.

Tian et al. introduced an enhanced YOLO-V3 model designed for real-time apple detection in orchards \cite{RN24}. Employing a camera with a 3000 x 4000-pixel resolution under varying weather and illumination conditions, the researchers gathered image data. Data augmentation techniques were applied to augment the dataset, enhancing its diversity. To optimize feature layers with low resolution in the YOLO-V3 network, the authors incorporated the DenseNet method, aiming to boost feature propagation, encourage reuse, and enhance overall network performance. The proposed YOLOV3-dense model exhibited superior performance compared to the original YOLO-V3 and the Faster R-CNN with VGG16 net model, particularly in terms of detection accuracy and real-time capabilities. Noteworthy, the study concentrated exclusively on apple detection, neglecting exploration into other fruits or crops. Future investigations could explore the adaptability of the proposed model to diverse agricultural contexts. Furthermore, the study overlooked the potential impact of adverse weather conditions, such as rain or fog, on the model's performance. Subsequent research endeavours could delve into assessing the robustness of the model under various weather scenarios. Lastly, the study's evaluation on a large-scale dataset was lacking. Consequently, forthcoming studies may consider validating the model on a more extensive dataset to further establish and validate its performance.

Sharpe and the research team developed a precision applicator for effective goosegrass control in Florida's vegetable plasticulture production \cite{RN26}. The study assessed the utilization of the YOLOv3-tiny detector for on-site goosegrass detection and spraying. Image processing involved various plants, including strawberry and tomato plants, as well as other weed species, to train and test the neural network. While showcasing the potential of convolutional neural networks in horticultural crop weed detection and management, the study identifies specific limitations and areas for improvement. Primarily, the focus on goosegrass detection in strawberry and tomato production warrants further research to extend the applicability of the network to diverse crops and weed species. The study highlights the superiority of the LB annotation method for production and precision spraying but suggests that additional classes or grouping may enhance overall network accuracy. Lastly, for tomatoes, the study acknowledges limitations in the piecewise image methodology, urging further research to enhance detection accuracy for this particular crop.

Junos et al. introduced YOLO-P, an object detection model capable of identifying and locating objects (FFB, grabber, and palm tree) in oil palm plantations \cite{RN27}. Through multiple experiments, the proposed model demonstrated outstanding mean average precision and F1 scores of 98.68\% and 0.97, respectively. Characterized by a faster training process and lightweight design (76 MB), the model exhibited accuracy in identifying fresh fruit bunches at various maturities, offering potential applications in automated crop harvesting systems. The comprehensive experimental results suggest that YOLO-P can accurately and robustly detect objects in palm oil plantations, thereby contributing to increased productivity and optimized operational costs in the agricultural industry.

Chen et al. made significant strides in citrus fruit detection with the development of the CitrusYOLO algorithm \cite{RN28}. Enhancements to the YOLOv4 model included the addition of a 152*152 feature detection layer, dense connections for multiscale fusion, and integration of depthwise separable convolution and attention mechanism modules. These improvements resulted in increased detection accuracy and real-time performance. CitrusYOLO exhibited superior performance, outperforming standard deep learning algorithms in terms of accuracy and time efficiency. Despite these advancements, the study recognizes certain limitations and areas for improvement. The dataset's focus on two varieties of Kumquats and Nanfeng tangerines and four varieties of Fertile orange, Tangerine, Mashui orange, and Gonggan suggests the potential for improved performance with expanded citrus varieties and growth stages. The algorithm's performance under varied lighting conditions and its applicability to other fruits or objects in different environments remain unknown. While experiments demonstrated effectiveness, real-world applications such as orchard yield estimation and fruit harvesting robots require further validation.

Hong et al. developed a lightweight model for detecting wheat ear Fusarium head blight (FHB) using RGB images \cite{RN29}. Leveraging YOLOv4 and MobileNet architectures, the proposed model struck a balance between accuracy and real-time FHB detection. With an accuracy of 93.69\% in detecting wheat ear FHB, outperforming the MobileNetv2-YOLOv4 model, the suggested model's reduced size facilitates deployment on uncrewed aerial vehicles (UAVs). While demonstrating great potential for real-time FHB detection, the study acknowledges certain limitations and areas for improvement. The study's exclusive focus on wheat ear Fusarium head blight detection, overlooking other diseases affecting wheat crops, suggests future research considerations for expanding the model's disease detection capabilities. Challenges related to incorrect detections of small objects and detection performance in complex backgrounds highlight the need for refining the model's performance. Additionally, the study's emphasis on reducing parameters necessitates exploration of the model's generalization across different edge platforms.

Wang et al. proposed a real-time method for vehicle identification and tracking using Improved YOLOv4 and binocular positioning (BPO) \cite{RN30}. Addressing the tracking problem in agricultural master-slave follow-up operations, the study's experiments showcased the method's accuracy in identifying and tracking the master vehicle in real time. Low RMS errors in terms of longitudinal, lateral, and heading angle deviation signify the method's effectiveness in meeting the positioning requirements of the master vehicle. The study also created a dataset for training and testing the identification model, hinting at future work on constructing and testing enslaved person automatic follow-up systems. While the study does not explicitly mention limitations, potential areas for further exploration include the validation of the method in diverse agricultural environments and consideration of its performance under varying weather and lighting conditions. Table \ref{tab:crop_detection} compiles a summary of research endeavours utilizing diverse YOLO architectures for crop detection within the agricultural domain.

\begin{table}[!ht]
\caption{Studies on Crop Detection Using YOLO}
\label{tab:crop_detection}
\centering
\resizebox{\columnwidth}{!}{%
\begin{tabular}{p{2cm}p{4cm}p{7cm}}
\hline
\textbf{Author} & \textbf{Model} & \textbf{Details} \\
\hline
\cite{RN24} & YOLOv3-dense & Enhanced YOLO-V3 model for real-time apple detection in orchards. Superior performance in detection accuracy and real-time capabilities. \\
\cite{RN26} & YOLOv3-tiny & Utilization of YOLOv3-tiny detector for on-site goosegrass detection and spraying in vegetable plasticulture production. Potential for convolutional neural networks in horticultural crop weed detection and management. \\
\cite{RN27} & YOLO-P & Introduction of YOLO-P for identifying and locating objects (FFB, grabber, and palm tree) in oil palm plantations. Outstanding mean average precision (98.68\%) and F1 score (0.97). \\
\cite{RN28} & Citrus-YOLO & Development of CitrusYOLO algorithm for citrus fruit detection. Improved YOLOv4 model with increased detection accuracy and real-time performance. \\
\cite{RN29} & YOLOv4 & Lightweight model for detecting wheat ear Fusarium head blight (FHB) using RGB images. YOLOv4 and MobileNet architectures for a balance between accuracy and real-time FHB detection. \\
\cite{RN30} & YOLOv & Real-time method for vehicle identification and tracking using Improved YOLO (IYO) v4 and binocular positioning (BPO). Accuracy in identifying and tracking master vehicles in agricultural master-slave follow-up operations. \\
\hline
\end{tabular}%
}
\end{table}

\subsection{Animal Tracking with YOLO}

In recent years, the application of You Only Look Once (YOLO) object detection models has revolutionized animal tracking in ecological research. YOLO's real-time and high-precision capabilities make it a compelling tool for monitoring wildlife, enabling the automatic identification and tracking of animals in diverse environments. This subsection delves into the innovative use of YOLO-based systems for animal tracking, exploring their contributions to understanding animal behaviour, migration patterns, and ecological dynamics.

Wang et al. conducted a study to monitor and analyze the behaviours of egg breeders in self-breeding cages using visual image processing and the YOLO v3 deep learning algorithm \cite{RN32}. Their approach identified six behaviours, achieving high precision rates and offering insights into the welfare state of egg breeders. Despite the success, the study's limitation involves the analysis of a limited sample size of Hy-Line Gray egg breeders in a single cage, warranting further investigation across diverse breeds, larger populations, and comparisons with existing methods for behaviour recognition \cite{RN32}.

Schütz et al. applied YOLOv4 for red fox detection and motion monitoring, showcasing the potential of computer vision systems in studying animal behaviour \cite{RN33}. While the study emphasized the efficiency and accuracy of computer evaluation, limitations included occasional camera manipulation by foxes and challenges in detecting rarely occurring fox positions not present in the training set. Proposed solutions involve securing the camera appropriately and expanding the training set to address bias and improve accuracy \cite{RN33}.

Yu et al. utilized the YOLO-improved model and edge computing to detect dairy cow feeding behaviour automatically \cite{RN35}. The proposed DRN-YOLO algorithm exhibited improved precision, recall, mean Average Precision (mAP), and F1-score, with potential areas for improvement outlined, such as further subdividing cow foraging behaviour and testing generalizability across diverse farm environments and cow populations \cite{RN35}.

Elmessery et al. conducted a comprehensive study to develop and validate a YOLOv8-based model for automatically detecting broiler pathological phenomena in intensive poultry houses \cite{RN36}. Despite successful training and detection, limitations included a constrained dataset of diseased broilers due to disease-related constraints and potential impacts of illumination intensity on image quality \cite{RN36}.

Barreiros et al. proposed the development of a tracking algorithm using YOLOv2 and Kalman filter to accurately track the movements of groups of zebrafish in a controlled experimental setup \cite{barreiros2021zebrafish}. They successfully implemented a system that could detect and track individual fish within a group, delimiting the region of fish heads for detection and estimating the best state of the fish's head position in each frame using the Kalman filter

Rančić et al. developed and tested a pipeline for detecting animals, specifically deer, using YOLOv3, YOLOv4, YOLOv4-tiny, and SSD applied to UAV images \cite{RN37}. While the study achieved high-performance predictions, limitations included the challenge of limited data addressed through pre-trained models, indicating the need for further research to enhance the system's robustness and generalizability \cite{RN37}.

Zheng et al. proposed the YOLO-BYTE algorithm for tracking multiple dairy cows using a single camera \cite{RN38}. Despite achieving high precision in dairy cow target detection, potential environmental impacts on accuracy were acknowledged, emphasizing the algorithm's need for further examination across diverse datasets and scenarios \cite{RN38}.

Wangli, et al. proposed a new pig detection and counting model called YOLOv5-SA-FC, which integrates shuffle attention and Focal-CIoU loss into the YOLOv5 framework backbone \cite{hao2023yolov5}. By leveraging the shuffle attention module, the model dynamically focuses on relevant features for pig detection and counting while reducing the weights of non-essential features. Additionally, the Focal-CIoU loss mechanism prioritizes predicted boxes with higher overlap with the target box, improving detection performance. Similarly, Jonggwan et al. developed an EmbeddedPigCount technology that utilizes  TinyYOLOv4 to accurately count pigs on large-scale pig farms \cite{kim2022embeddedpigcount}. They collected image data from a commercial pig farm in Korea, where a surveillance camera captured images of pigs and humans moving in a hallway. The researchers manually annotated bounding boxes in the images and used a total of 2675 images for training the detection module. The research achieved a counting accuracy of 99.44\% even when pigs passed through the counting zone back and forth.

\subsection{Disease Detection in Agriculture Using YOLO}

\begin{figure*}[h]
\centering
\includegraphics[width=1\columnwidth]{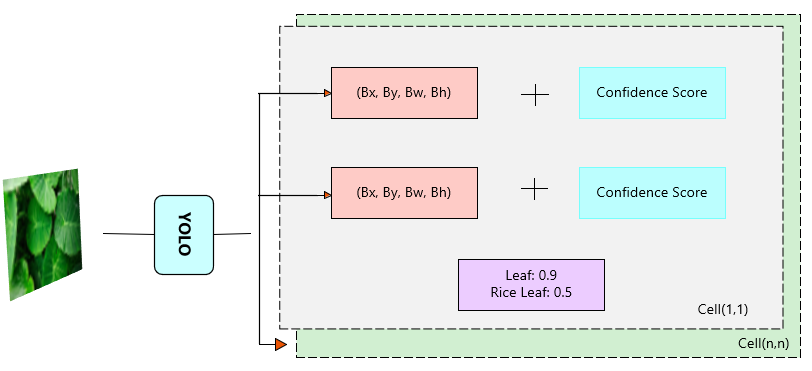}
\caption{Real-time YOLO-based Leaf Disease Detection}
\label{Figure:8}
\end{figure*}

The agricultural sector faces constant challenges in maintaining crop health and ensuring optimal yields. Identifying and addressing plant diseases promptly are essential components of sustainable farming practices. With advancements in computer vision and deep learning, particularly the You Only Look Once (YOLO) algorithm, there has been a growing interest in leveraging these technologies for automated disease detection in crops.

This subsection explores several studies that employ YOLO-based models to detect and monitor diseases in agricultural settings. These studies showcase the potential of YOLO in providing accurate and efficient solutions for disease identification, contributing to the development of intelligent and technology-driven approaches in modern agriculture.

% Liu and Wang: Tomato Disease and Pest Detection with Improved YOLO v3
Liu and Wang conducted a study focusing on the detection of tomato diseases and insect pests in natural environments, culminating in the creation of a dedicated dataset \cite{RN12}. Utilizing the YOLO v3 model, they achieved a commendable detection accuracy of 92.39\% in a swift 20.39 ms. The improved YOLO v3 surpassed alternative methods in both accuracy and speed, including SSD, Faster R-CNN, and the original YOLO v3. While the algorithm demonstrated effectiveness in real-time detection, there remains an opportunity to enhance both accuracy and speed for practical applications, particularly in comparison to the high precision attainable through deep learning-based classification methods.

% Morbekar et al.: YOLO-based Crop Disease Detection Model
Morbekar and colleagues developed a real-time crop disease detection model utilizing the YOLO object detection method \cite{RN13} for detection and classification, as shown in Figure \ref{Figure:8}. The model exhibited promising results with a notable accuracy of 98.5\% when tested on the PlantVillage dataset. However, the study's limitations include the dataset's focus on major crops in India, potentially limiting its applicability to other crops and regions. Additionally, the system's scope is confined to detecting diseases solely on leaves, neglecting other crucial parts of the crop, such as stems or fruits.

% Nihar and Raghavendra: YOLO Algorithm for Real-Time Rice Crop Disease Detection
Nihar and Raghavendra proposed an innovative approach for real-time rice crop disease detection, developing a state-of-the-art deep learning model based on the tiny\_yolov3 algorithm \cite{RN14}. Achieving an impressive accuracy rate of 98.92\%, their model facilitates early identification of potential issues, allowing farmers to proactively safeguard their crops. The authors suggested the model's adaptability for pest detection, enhancing its versatility for diverse applications.

% Agbulos et al.: YOLO Algorithm for Rice Leaf Disease Identification
Agbulos et al. employed the YOLO Algorithm to identify rice leaf diseases, achieving an overall accuracy of 73.33\% \cite{RN15}. While successful in identifying leaf blast and brown spot diseases, the study's focus on static rice leaf images and the hardware setup's limitations, including the Raspberry Pi 3 and camera module, pose potential challenges in real-world scenarios. Future improvements may involve upgrading hardware components for enhanced image quality and exploring a broader spectrum of diseases in rice plants.

% Lippi et al.: YOLO-based Insect Detection in Hazelnut Orchards
Lippi and colleagues developed an insect detection system using a YOLO-based Convolutional Neural Network (CNN) for identifying true bugs in hazelnut orchards \cite{RN16}. With an average precision of approximately 94.5\%, the system showcased real-time processing capabilities. However, scalability concerns arise when applying the system to large orchards, and potential challenges associated with the system's depth sensor resolution may impact its performance.

\begin{table*}[!ht]
\caption{Studies on Crop Disease and Pest Detection Using YOLO}
\label{tab:crop_disease_pest_detection}
\centering
\resizebox{\textwidth}{!}{%
\begin{tabular}{p{1.5cm}p{4.5cm}p{6.5cm}}
\hline
\textbf{Authors} & \textbf{Model} & \textbf{Details} \\
\hline
\cite{RN12} & YOLO v3 & Detection of tomato diseases and insect pests. Accuracy of 92.39\% in 20.39 ms. \\
\cite{RN13} & YOLO & Real-time crop disease detection. Accuracy of 98.5\% on PlantVillage dataset. \\
\cite{RN14} & YOLO-v3 Tiny & Real-time rice crop disease detection. Accuracy rate of 98.92\%. \\
\cite{RN15} & YOLO & Identification of rice leaf diseases. Overall accuracy of 73.33\%. \\
\cite{RN16} & YOLO-based CNN & Insect detection in hazelnut orchards. Average precision of approximately 94.5\%. \\
\cite{RN17} & YOLOv4 & Mulberry crop disease detection. High speed and accuracy. \\
\cite{RN18} & YOLO V3 & Apple tree disease detection. Improved accuracy and faster results. \\
\cite{RN19} & YOLO & Insect detection in soybean crops. High mean average precision. \\
\cite{RN20} & YOLO v5 & Seed classification in agriculture. Precision and recall of 99\%. \\
\cite{RN21} & YOLO v5 & Bell pepper plant disease identification. Superior accuracy and reduced model size. \\
\cite{RN22} & YOLOv7 & Tea leaf disease detection. AI-based solution for Bangladesh's tea cultivation. \\
\cite{RN23} & YOLO-Tea model & Improved model for tea disease and insect pest detection. \\
\hline
\end{tabular}%
}
\end{table*}

% Reddy and Deeksha: YOLOv4 Model for Mulberry Crop Disease Detection
Reddy and Deeksha trained a YOLOv4 model for detecting and identifying leaf diseases in mulberry crops, achieving high speed and accuracy \cite{RN17}. The model's capability to recommend corresponding pesticides post-detection holds promise for effective disease management. The study, while successful in its focus on mulberry crops, encourages further exploration for identifying various diseases and considering real-time video classification.

% Mathew and Mahesh: YOLO V3 Networks for Apple Tree Disease Detection
Mathew and Mahesh explored the significance of early disease detection in apple trees, employing YOLO V3 networks for disease detection \cite{RN18}. The study highlighted YOLO V3's benefits, such as faster results and improved accuracy, but acknowledged challenges, including environmental interference and the need for continuous plant health monitoring.

% Verma et al.: YOLO Algorithms for Insect Detection in Soybean Crops
Verma et al. proposed a framework employing YOLO algorithms for insect detection in soybean crops \cite{RN19}. While achieving high mean average precision, the study acknowledged limitations, including dataset size constraints and occasional misclassifications. The potential ethical and environmental implications of pesticide application based on insect detection were also noted for further consideration.

% Kundu et al.: YOLO v5 for Seed Classification in Agriculture
Kundu et al. developed a YOLO v5-based system for automated seed segregation and classification, achieving high precision and recall of 99\% \cite{RN20}. While impactful in classifying pearl millet and maize seeds, the study identified the need for further research in seed classification for mixed cropping scenarios and the broader categorization of seeds based on crop type and quality.

% Mathew and Mahesh: YOLO v5 for Bell Pepper Plant Disease Identification
Mathew and Mahesh employed YOLO v5 for identifying diseases in bell pepper plants, demonstrating superior accuracy and reduced model size \cite{RN21}. The study hinted at the potential extension of disease detection to various diseases affecting bell pepper plants, promising improvements in farm yield through prompt disease identification.

% Soeb et al.: YOLOv7 for Tea Leaf Disease Detection
Soeb et al. introduced an AI-based solution using the YOLOv7 approach for detecting tea leaf diseases, emphasizing the need to explore AI's benefits in Bangladesh's tea cultivation \cite{RN22}. Acknowledging limitations such as limited labelled data and a lack of established evaluation metrics, the study advocated for further research to enhance the YOLOv7 model's effectiveness in tea leaf disease detection.

% Xue et al.: Improved YOLO-Tea Model for Tea Disease Detection
Xue et al. proposed the YOLO-Tea model, addressing small target challenges for tea diseases and insect pests with improved feature extraction and attention mechanisms \cite{RN23}. While showcasing its potential through ablative experiments and comparisons, the study emphasized the need for continued exploration and evaluation, particularly in real-world tea disease monitoring applications. Table \ref{tab:crop_disease_pest_detection} provides an overview of studies employing various YOLO architectures for the detection of crop diseases and pests in the agricultural domain.

% Precision Framing Using YOLO: Introduction
\subsection{Precision Framing Using YOLO}

Precision farming, a crucial aspect in various agricultural applications, involves accurately delineating and identifying specific targets within an image or video frame. The application of You Only Look Once (YOLO) algorithms in precision framing has demonstrated remarkable capabilities in detecting and classifying objects with speed and efficiency. This subsection explores several studies that leverage YOLO for precision framing in diverse agricultural contexts, showcasing its effectiveness in tasks such as disease and pest detection, seed classification, and more. These applications not only enhance the accuracy of target identification but also contribute to optimizing agricultural processes and improving overall crop management.

% Li et al. - Agricultural Greenhouse Detection
Li et al. employed convolutional neural networks (CNNs) to identify agricultural greenhouses (AGs) in high-resolution satellite images \cite{RN39}. The study compared the performance of three prominent CNN-based object detection models: Faster R-CNN, YOLO v3, and SSD. Utilizing the PyTorch deep learning framework on a workstation equipped with two TITAN RTX GPUs, the authors trained and evaluated the models. By fusing GF-1 data into 2 m multispectral data along with GF-2, the study enhanced sample diversity and assessed method transferability across distinct data sources exhibiting similar AG styles. The findings indicated that the YOLO v3 model surpassed the other models in terms of accuracy and efficiency for AG detection. While the research contributes valuable insights into AG detection methodologies and CNN-based object detection in remote sensing, the authors recognized the need for future investigations to leverage multispectral and hyperspectral data in satellite images for improved object detection. The study's exclusive focus on AG detection prompts further exploration of other geospatial objects in future research.

% Khan et al. - Crop and Weed Differentiation
Khan et al. developed a deep learning system for distinguishing between crops and weeds in strawberry and pea fields, aiming for integration into precision agriculture sprayers for real-time weed management \cite{RN40}. Unmanned aerial vehicles equipped with cameras captured field images, and the deep learning techniques were optimized for high accuracy in identifying small weed patches, particularly in early growth stages. The system exhibited an overall average accuracy of 94.73\%, outperforming existing machine learning and deep learning-based approaches. Despite its robustness, the study acknowledged limitations related to dataset size, generalizability, and real-time integration into precision spraying systems, calling for additional research and development.

% Mamdouh and Khattab - Olive Fruit Fly Detection
Mamdouh and Khattab introduced a YOLO-based deep learning framework for detecting and counting olive fruit flies in orchards \cite{RN41}. The framework demonstrated exceptional precision (0.84), recall (0.97), F1-score (0.9), and mean Average Precision (mAP) of 96.68\%, surpassing existing pest detection systems. The authors highlighted the framework's potential benefits over traditional manual methods, emphasizing its effectiveness through extensive simulation experiments. Recognizing the framework's limitations, including a lack of a large-scale dataset, the authors proposed future improvements such as real-life image evaluation, dataset enrichment, and transformation into a multi-class classifier.

\subsection{Comparative Analysis}
Our methodology systematically compares different YOLO applications within agricultural settings, integrating key aspects such as task complexity, control over experimental conditions, hardware dependencies, and a critical analysis of results and error measurement methods. This holistic approach not only assesses performance but also elucidates the practical implications of deploying these models in real-world agricultural scenarios.

In the comparative analysis of different YOLO versions across various application domains, particularly in agriculture, each version of the YOLO architecture demonstrates unique strengths and encounters specific limitations that influence its suitability for certain tasks (see Table \ref{tab:yolo_comparative_analysis}). For instance, YOLOv1 primarily addressed standard datasets like VOC 2007 or Sesame
Fields Image Dataset with straightforward object recognition tasks like weed detection. In contrast, later versions like YOLOv4 and YOLOv5 have been utilized in more complex agricultural datasets that feature varied backgrounds, multiple object classes, and demand real-time detection capabilities. These tasks are evaluated not only for accuracy but also for the model's ability to manage the intricacies of natural scenes, including variable lighting conditions, occlusions, and overlapping objects.

Our analysis underscores the importance of controlling experimental conditions to validate the robustness of object detection models. YOLOv3's deployment in UAV-based weed detection, for instance, involves not just the algorithm's performance but also factors like flight stability, camera quality, and environmental interference—all of which significantly impact outcomes. Similarly, YOLOv6's application in wildlife monitoring presents challenges such as varying animal speeds and camouflaged backgrounds, pushing the limits of detection capabilities under less controlled but highly variable conditions. The choice of hardware significantly influences the deployment of YOLO models. Our review spans hardware from high-end GPUs for training to embedded systems like NVIDIA Jetson for infield deployment, critically evaluating performance metrics like frame rate, processing speed, and power consumption to determine feasibility in agricultural settings where limited power and mobility often dictate hardware choices.

Our methodology includes a critical review of error measurement techniques, now detailed in the 'Performance Metrics' column of our comparative analysis table (see Table \ref{tab:yolo_comparative_analysis}). We focus on metrics such as precision, recall, mAP, and F1 score, which are crucial for assessing model performance across different scenarios. Additionally, we categorize errors into types such as localization, classification, and false positives/negatives. This categorization not only provides a nuanced view of model performance but also enhances our understanding of the practical implications of deploying these models in field conditions. For instance, in YOLOv4's application in crop disease detection, understanding the impact of false negatives is crucial, as missing a diseased plant could lead to wider spread within the crop.

Compiling data from this comprehensive methodology allows us to synthesize findings across different YOLO applications and versions, highlighting trends like improvements in speed and accuracy versus trade-offs in computational demand and complexity. This synthesis not only addresses current research questions but also identifies gaps for future research, suggesting potential model training or deployment enhancements to better meet the specific needs of precision agriculture. The evolution of the YOLO architectures marks a significant trajectory of technological enhancements, methodically refined to address the diverse and challenging demands of agricultural applications. Each version from YOLOv1 through YOLOv10 has been adapted to overcome specific limitations, leading to more sophisticated systems capable of complex environmental interactions. These iterative advancements highlight the importance of selecting the appropriate YOLO variant tailored to specific agricultural tasks, balancing computational demands with precision requirements for tasks such as pest identification and crop disease monitoring. By aligning the model capabilities with task-specific needs, researchers and practitioners can harness YOLO technologies to propel the future of precision agriculture and sustainable farming practices effectively.

This comparative analysis highlights the critical role of model selection based on specific application needs and the technological trade-offs involved, providing a framework for future research and application in technology-driven agriculture.

\begin{table}[h]
\centering
\caption{Comparative Analysis of Different YOLO Versions Across Various Application Domains}
\label{tab:yolo_comparative_analysis}
\begin{tabularx}{\textwidth}{>{\raggedright\arraybackslash}p{2cm}>{\raggedright\arraybackslash}X>{\raggedright\arraybackslash}X>{\raggedright\arraybackslash}X}
\toprule
\textbf{YOLO Version} & \textbf{Application Domain} & \textbf{Strengths} & \textbf{Limitations} \\
\midrule
YOLOv1 & Object Detection, Weed and Crop Detection \cite{RN3, RN8, RN9, RN30, RN13, RN15, RN18, RN19} & Fast real-time processing & Struggles with small objects \\
YOLOv2 & Animal Tracking \cite{barreiros2021zebrafish} & Improved recall, handles small objects better & Higher computational requirements \\
YOLOv3 & Agriculture \cite{RN1,RN10, RN24, RN26, RN32, RN12, RN14, RN18}  & Multi-scale detection, good in diverse conditions & Not optimized for very low-power devices \\
YOLOv4  & Agriculture \cite{RN2, RN6, RN29, RN33, RN17, kim2022embeddedpigcount} & Robust in complex visual environments & Setup complexity for training on custom datasets\\
YOLOv5  & Weed Detection \cite{RN5, RN7, RN20, RN21, hao2023yolov5} & Very fast, suitable for real-time applications & May require fine-tuning for specific scenarios \\
YOLOv6 & Wildlife Monitoring \cite{mao2024improved} &  High accuracy with enhanced depth & Needs high computation power for best results \\
YOLOv7 & Crop Disease Detection \cite{RN22}  & High precision, effective in crowded scenes & Computational efficiency decreases with scale \\
YOLOv8 & Agriculture \cite{RN36, shi2023multi} & Very high speed, suitable for dynamic environments & Can struggle with very small or fast-moving objects \\
YOLOv9 & Plant Disease detection \cite{chien2024yolov9, boudaa2024advancing} & High accuracy and recall, suitable for detailed medical scans & Requires extensive dataset for training \\
YOLOv10 & --- & Enhanced accuracy-efficiency trade-off, NMS-free model & Complex configuration for multi-source integration \\
\bottomrule
\end{tabularx}
\end{table}

% End of Introduction to Disease Detection in Agriculture Using YOLO

\section{Discussion}

The integration of YOLO variants in agriculture has emerged as a transformative approach, revolutionizing various aspects of farming and crop management. As evidenced in Table \ref{tab:weed_detection}, \ref{tab:crop_detection} and \ref{tab:crop_disease_pest_detection}, the diverse range of YOLO models employed for agricultural applications showcases the adaptability and efficacy of this architecture in addressing the unique challenges within the agricultural domain. Notably, the consistently high levels of accuracy achieved across different applications, including weed detection, crop identification, and disease diagnosis, underscore the robust performance of YOLO-based models in diverse agricultural scenarios.

\textbf{Real-time Precision Agriculture:}
One of the standout features of YOLO variants in agriculture is their ability to facilitate real-time precision farming. The models, such as YOLOv3, YOLOv4, and YOLOv5, have demonstrated exceptional swiftness and accuracy in detecting and identifying objects in agricultural landscapes. This real-time capability holds significant promise for optimizing farming practices, enabling timely decision-making, and enhancing resource allocation efficiency.

\textbf{Weed Detection and Management:}
The application of YOLO variants in weed detection, as illustrated by studies such as~\cite{RN1, RN2, RN3, RN4, RN5, RN6, RN7, RN8, RN9, RN10}, signifies a paradigm shift in traditional weed management practices. The real-time identification and localization of weeds using YOLO models empower farmers to implement targeted and efficient weed control measures. Despite certain challenges, such as resolution limitations in detecting smaller objects, the overall performance of YOLO-based systems like YOLO-WEED~\cite{RN1} and YOLOv4-weeds~\cite{RN2} indicates their potential for widespread adoption in weed management.

\textbf{Crop Detection and Monitoring:}
The significance of robust crop detection methodologies in precision agriculture cannot be overstated. YOLO variants, as exemplified by studies such as~\cite{RN24, RN26, RN27, RN28, RN29, RN30}, contribute to accurate identification and delineation of crops in large-scale fields. These models offer a comprehensive solution for monitoring crop health, estimating yields, and optimizing resource management. The development of specialized models like CitrusYOLO~\cite{RN28} and YOLO-P~\cite{RN27} for specific crops further emphasizes the adaptability of YOLO architectures to diverse agricultural settings.

\section{Challenges in YOLO-based Agricultural Applications}
While the achievements in agricultural applications of YOLO are remarkable, challenges persist. The regional specificity in some datasets, hardware constraints, and the need for further research to enhance model applicability are areas that warrant attention. Future work should focus on addressing these challenges, exploring the generalizability of models to different crops and environmental conditions, and fostering advancements in sensing technologies for broader applications.

\subsection{Data Specificity and Generalization}
One notable challenge in YOLO-based agricultural applications lies in the specificity of datasets used for model training. Many studies focus on particular crops or regions, which may limit the model's generalizability across diverse agricultural landscapes~\cite{kiala2022determining}. Addressing this challenge involves creating more comprehensive and diverse datasets that encompass various crops, growth stages, and environmental conditions~\cite{su2020advanced}. Additionally, research efforts should aim at developing transfer learning techniques to enhance model adaptability to new agricultural contexts~\cite{shendryk2020leveraging}.

\subsection{Hardware Limitations}
The deployment of YOLO-based systems in real-world agricultural settings may encounter hardware constraints, especially in resource-limited environments~\cite{zahid2023lightweight}. Many studies leverage powerful computing resources for model training and inference, but practical implementation on edge devices or embedded systems poses challenges~\cite{hussain2023exudate}. Future research should explore model optimization techniques, quantization, and lightweight architectures to make YOLO variants more accessible for deployment on edge devices commonly used in precision farming equipment.

\subsection{Environmental Variability}
Agricultural environments are inherently dynamic, with variations in lighting conditions, weather, and terrain. YOLO models, while robust, may face challenges in adapting to these environmental changes. Ensuring the reliability of detection under diverse conditions requires the development of models that are resilient to variations in illumination, adverse weather, and different terrains. This necessitates the incorporation of environmental adaptability in model training and further exploration of techniques like domain adaptation.

\subsection{Small Object Detection}
Identifying diminutive weeds or diseases in agricultural contexts presents a unique set of hurdles for YOLO variants. The intrinsic structure of YOLO may encounter difficulties in discerning smaller objects within an image. Addressing this obstacle necessitates advancements in feature extraction, attention mechanisms, or the exploration of multi-scale detection strategies. Moreover, ensemble techniques can be used to overcome false positives and negatives as done by researchers \cite{ahn2021ensemblepigdet, li2023high, lee2022detecting, singh2023object}. It is crucial for future research to concentrate on refining YOLO architectures to augment the precision of detecting small objects in precision farming applications. Furthermore, the integration of attention mechanisms can serve as a valuable approach, steering YOLO architectures towards subtle defects, as has been successfully implemented in industries like textiles~\cite{jin2021automatic,zhang2022improved, article-53, inproceedings-58,article-62,rong2021fabric}.

\section{Future Directions and Opportunities}

\subsection{Multi-Modal Integration}
The integration of multi-modal data sources, such as combining RGB images with thermal or hyperspectral data, holds great potential for advancing YOLO-based agricultural applications~\cite{amigo2019preprocessing}. Combining different modalities can provide richer information for more accurate and robust detection of crops, weeds, and diseases. Future research should explore the fusion of diverse data sources to enhance the overall performance and reliability of YOLO models in precision agriculture.

\subsection{Explainability and Interpretability}
As YOLO models become integral to decision-making in agriculture, ensuring their explainability and interpretability is crucial. Farmers and stakeholders need to understand the rationale behind model predictions to trust and effectively implement precision farming practices. Future work should focus on developing methodologies for explaining YOLO model decisions and providing insights into how and why certain detections are made, especially in complex and dynamic agricultural environments.

\subsection{Real-Time Adaptive Systems}
The evolution of YOLO architectures toward real-time capabilities opens avenues for developing adaptive systems that respond dynamically to changing agricultural conditions. Future YOLO-based models could incorporate real-time learning mechanisms, enabling them to adapt and improve their performance over time based on continuous feedback from the field. This would contribute to the development of intelligent and self-improving precision agriculture systems.

\subsection{Human-AI Collaboration}
Recognizing the expertise of farmers, future research should explore models that facilitate human-AI collaboration in decision-making processes. Integrating farmer knowledge with AI-driven insights can lead to more effective and context-aware agricultural practices. Human-AI collaboration is vital for addressing the complexities and uncertainties inherent in agriculture, allowing for seamless integration of YOLO-based technologies into the existing farming ecosystem.

In summary, overcoming the challenges and leveraging future opportunities requires a concerted effort from the research community, industry stakeholders, and farmers. The continuous refinement of YOLO-based models, coupled with advancements in data collection, hardware, and interpretability, will propel the application of AI in agriculture towards sustainable and efficient farming practices.

\section{Conclusion}

In conclusion, the intersection of YOLO variants and agriculture presents a transformative potential for precision farming, weed management, and crop monitoring. The consistent advancements and promising results showcased in various studies underscore the pivotal role of YOLO architectures in shaping the future of smart and efficient agriculture. The comprehensive review of YOLO variants in agricultural applications highlights the transformative potential of these models in revolutionizing precision farming. From crop detection and disease identification to weed management, YOLO variants have showcased remarkable capabilities, offering real-time and efficient solutions to long-standing challenges in agriculture. The discussion and analysis of various studies underscore the versatility and adaptability of YOLO architectures across diverse agricultural scenarios. Despite the evident successes, challenges persist, necessitating ongoing research efforts. The specificity of training datasets, hardware limitations, and environmental variability pose hurdles that demand innovative solutions. Future research should prioritize the development of more inclusive datasets, optimization techniques for edge devices, and models resilient to dynamic agricultural environments.

Looking ahead, multi-modal integration, explainability, and real-time adaptive systems present exciting opportunities for further enhancing the utility of YOLO models in agriculture. The fusion of different data modalities, coupled with real-time learning mechanisms, can usher in a new era of intelligent and context-aware precision farming. Additionally, a focus on human-AI collaboration acknowledges the indispensable role of farmers in the decision-making process, promoting a harmonious integration of AI technologies into existing agricultural practices. In conclusion, the evolution of YOLO variants in agriculture signifies a paradigm shift towards sustainable, efficient, and technologically-driven farming practices. As researchers, practitioners, and stakeholders collaborate, the future holds great promise for the continued advancement of YOLO-based applications, contributing to the global effort to address food security and promote environmentally conscious agriculture.

%%%%%%%%%%%%%%%%%%%%%%%%%%%%%%%%%%%%%%%%%%
\vspace{6pt} 

%%%%%%%%%%%%%%%%%%%%%%%%%%%%%%%%%%%%%%%%%%
%% optional
%\supplementary{The following supporting information can be downloaded at:  \linksupplementary{s1}, Figure S1: title; Table S1: title; Video S1: title.}


%\begin{tabularx}{\textwidth}{lll}
%\toprule
%\textbf{Name} & \textbf{Type} & \textbf{Description} \\
%\midrule
%S1 & Python script (.py) & Script of python source code used in XX \\
%S2 & Text (.txt) & Script of modelling code used to make Figure X \\
%S3 & Text (.txt) & Raw data from experiment X \\
%S4 & Video (.mp4) & Video demonstrating the hardware in use \\
%\bottomrule
%\end{tabularx}
%}



\bibliographystyle{unsrt}  % Changes bibliography style to unsorted
\bibliography{ref}  % This points to the filename of your BibTeX file without the .bib extension

\begin{thebibliography}{100}

\bibitem{ariana2006near}
Diwan~P Ariana, Renfu Lu, and Daniel~E Guyer.
\newblock Near-infrared hyperspectral reflectance imaging for detection of bruises on pickling cucumbers.
\newblock {\em Computers and electronics in agriculture}, 53(1):60--70, 2006.

\bibitem{hussain2023yolo}
Muhammad Hussain.
\newblock Yolo-v1 to yolo-v8, the rise of yolo and its complementary nature toward digital manufacturing and industrial defect detection.
\newblock {\em Machines}, 11(7):677, 2023.

\bibitem{zhu2019rapid}
Xiaolin Zhu and Guanghui Li.
\newblock Rapid detection and visualization of slight bruise on apples using hyperspectral imaging.
\newblock {\em International journal of food properties}, 22(1):1709--1719, 2019.

\bibitem{zaji2023wheat}
Amirhossein Zaji, Zheng Liu, Gaozhi Xiao, Pankaj Bhowmik, Jatinder~S Sangha, and Yuefeng Ruan.
\newblock Wheat spikes height estimation using stereo cameras.
\newblock {\em IEEE Transactions on AgriFood Electronics}, 2023.

\bibitem{mao2010confirmation}
Shuang-Lin Mao, Yu-Ming Wei, Wenguang Cao, Xiu-Jin Lan, Ma~Yu, Zheng-Mao Chen, Guo-Yue Chen, and You-Liang Zheng.
\newblock Confirmation of the relationship between plant height and fusarium head blight resistance in wheat (triticum aestivum l.) by qtl meta-analysis.
\newblock {\em Euphytica}, 174:343--356, 2010.

\bibitem{wang2018field}
Xu~Wang, Daljit Singh, Sandeep Marla, Geoffrey Morris, and Jesse Poland.
\newblock Field-based high-throughput phenotyping of plant height in sorghum using different sensing technologies.
\newblock {\em Plant Methods}, 14(1):1--16, 2018.

\bibitem{20}
S.~Liao, J.~Wang, R.~Yu, K.~Sato, and Z.~Cheng.
\newblock Cnn for situations understanding based on sentiment analysis of twitter data.
\newblock {\em \textit{Procedia Computer Science}}, 111:376--381, 2017.

\bibitem{21}
H.~Sak, A.~Senior, K.~Rao, and F.~Beaufays.
\newblock Fast and accurate recurrent neural network acoustic models for speech recognition.
\newblock {\em arXiv:1507.06947 [cs, stat]}, Jul. 2015.

\bibitem{22}
Mujadded Al~Rabbani~Alif, Sabbir Ahmed, and Muhammad~Abul Hasan.
\newblock Isolated bangla handwritten character recognition with convolutional neural network.
\newblock pages 1--6, 2017.

\bibitem{23}
Y.~Zhang, Y.~Tong, and Y.~Jiang.
\newblock Study of sentiment classification for chinese microblog based on recurrent neural network.
\newblock {\em Chinese Journal of Electronics}, 25(4):601--607, Jul. 2016.

\bibitem{24}
S.~Lai, L.~Xu, K.~Liu, and J.~Zhao.
\newblock Recurrent convolutional neural networks for text classification.
\newblock {\em Proceedings of the AAAI Conference on Artificial Intelligence}, 29(1), Feb. 2015.

\bibitem{25}
D.~Wei and et~al.
\newblock Research on unstructured text data mining and fault classification based on rnn-lstm with malfunction inspection report.
\newblock {\em Energies}, 10(3):406, Mar. 2017.

\bibitem{26}
D.~Quang and X.~Xie.
\newblock Danq: A hybrid convolutional and recurrent deep neural network for quantifying the function of dna sequences.
\newblock {\em Nucleic Acids Research}, 44(11):e107--e107, Apr. 2016.

\bibitem{boltvision}
Mujadded Al~Rabbani Alif, Muhammad Hussain, Gareth Tucker, and Simon Iwnicki.
\newblock Boltvision: A comparative analysis of cnn, cct, and vit in achieving high accuracy for missing bolt classification in train components.
\newblock {\em Machines}, 12(2):93, 2024.

\bibitem{alif2024lightweight}
Mujadded Al~Rabbani Alif and Muhammad Hussain.
\newblock Lightweight convolutional network with integrated attention mechanism for missing bolt detection in railways.
\newblock {\em Metrology}, 4(2):254--278, 2024.

\bibitem{attention}
Mujadded Al~Rabbani Alif.
\newblock Attention-based automated pallet racking damage detection.
\newblock 9(1), 2024.

\bibitem{27}
M.~R. Mezaal, B.~Pradhan, M.~I. Sameen, H.~Z.~Mohd Shafri, and Z.~M. Yusoff.
\newblock Optimized neural architecture for automatic landslide detection from high-resolution airborne laser scanning data.
\newblock {\em Applied Sciences}, 7(7):730, Jul. 2017.

\bibitem{28}
N.~Xu and et~al.
\newblock Dual-stream recurrent neural network for video captioning.
\newblock {\em IEEE Transactions on Circuits and Systems for Video Technology}, 29(8):2482--2493, Aug. 2019.

\bibitem{29}
J.~Kim, J.~Kim, H.~L. Thu, and H.~Kim.
\newblock Long short term memory recurrent neural network classifier for intrusion detection.
\newblock In {\em 2016 International Conference on Platform Technology and Service (PlatCon)}, 2016.

\bibitem{30}
A.~M. Rather, A.~Agarwal, and V.~N. Sastry.
\newblock Recurrent neural network and a hybrid model for prediction of stock returns.
\newblock {\em Expert Systems with Applications}, 42(6):3234--3241, Apr. 2015.

\bibitem{31}
Ming Liang and Xiaolin Hu.
\newblock Recurrent convolutional neural network for object recognition.
\newblock {\em IEEE Xplore}, Jun. 2015.
\newblock (accessed Dec. 03, 2020).

\bibitem{32}
W.~Nash, T.~Drummond, and N.~Birbilis.
\newblock A review of deep learning in the study of materials degradation.
\newblock {\em npj Materials Degradation}, 2(1), Nov. 2018.

\bibitem{16}
T.~Diwan, G.~Anirudh, and J.~V. Tembhurne.
\newblock Object detection using yolo: Challenges, architectural successors, datasets and applications.
\newblock {\em \textit{Multimedia Tools and Applications}}, Aug. 2022.

\bibitem{33}
Y.~Bengio, A.~Courville, and P.~Vincent.
\newblock Unsupervised feature learning and deep learning: A review and new perspectives, 2012.
\newblock Accessed: May 17, 2023.

\bibitem{34}
Ujjwal.
\newblock An intuitive explanation of convolutional neural networks.
\newblock \textit{the data science blog}, May 29 2017.

\bibitem{35}
K.~Chen, K.~Franko, and R.~Sang.
\newblock Structured model pruning of convolutional networks on tensor processing units.
\newblock arXiv:2107.04191 [cs], Jul. 2021.

\bibitem{36}
S.~Agarwal, J.~O.~D. Terrail, and F.~Jurie.
\newblock Recent advances in object detection in the age of deep convolutional neural networks.
\newblock arXiv.org, Aug. 20 2019.

\bibitem{37}
L.~Liu and et~al.
\newblock Deep learning for generic object detection: A survey.
\newblock Sep. 2018.

\bibitem{38}
C.-Y. Wang, H.-Y.~M. Liao, Y.-H. Wu, P.-Y. Chen, J.-W. Hsieh, and I.-H. Yeh.
\newblock Cspnet: A new backbone that can enhance learning capability of cnn.
\newblock openaccess.thecvf.com, 2020.
\newblock (accessed Apr. 23, 2023).

\bibitem{xie2021oriented}
Xingxing Xie, Gong Cheng, Jiabao Wang, Xiwen Yao, and Junwei Han.
\newblock Oriented r-cnn for object detection.
\newblock In {\em Proceedings of the IEEE/CVF international conference on computer vision}, pages 3520--3529, 2021.

\bibitem{girshick2015fast}
R.~Girshick.
\newblock Fast r-cnn.
\newblock In {\em International Conference on Computer Vision}, pages 1137--1149, 2015.

\bibitem{ren2017faster}
S.~Ren and et~al.
\newblock Faster r-cnn: towards real-time object detection with region proposal networks.
\newblock In {\em IEEE Transactions on Pattern Analysis and Machine Intelligence}, pages 1137--1149, 2017.

\bibitem{lin2017feature}
T.-Y. Lin and et~al.
\newblock Feature pyramid networks for object detection.
\newblock In {\em Proceedings of the 2017 IEEE Conference on Computer Vision and Pattern Recognition (CVPR)}, pages 936--944, 2017.

\bibitem{liu2016ssd}
W.~Liu and et~al.
\newblock Ssd: single shot multibox detector.
\newblock In {\em European Conference on Computer Vision}, 2016.

\bibitem{sun2020dense}
Chang Sun, Yibo Ai, Sheng Wang, and Weidong Zhang.
\newblock Dense-refinedet for traffic sign detection and classification.
\newblock {\em Sensors}, 20(22):6570, 2020.

\bibitem{39}
T.-Y. Lin, P.~Goyal, R.~Girshick, K.~He, and Piotr Dollár.
\newblock Focal loss for dense object detection.
\newblock {\em arXiv (Cornell University)}, Aug. 2017.

\bibitem{17}
J.~Redmon, S.~Divvala, R.~Girshick, and A.~Farhadi.
\newblock You only look once: Unified, real-time object detection.
\newblock In {\em CVPR}, pages 779--788, 2016.

\bibitem{61}
J.~Redmon and A.~Farhadi.
\newblock Yolo9000: Better, faster, stronger.
\newblock Dec. 2016.

\bibitem{Borisyuk2018}
F.~Borisyuk, A.~Gordo, and V.~Sivakumar.
\newblock Rosetta.
\newblock In {\em Proceedings of the 24th ACM SIGKDD International Conference on Knowledge Discovery \& Data Mining}, Jul. 2018.

\bibitem{Won2019}
J.-H. Won, D.-H. Lee, K.-M. Lee, and C.-H. Lin.
\newblock An improved yolov3-based neural network for de-identification technology.
\newblock {\em IEEE Xplore}, Jun. 2019.

\bibitem{Everingham2010}
M.~Everingham, S.~M.~A. Eslami, L.~Van~Gool, C.~K.~I. Williams, J.~Winn, and A.~Zisserman.
\newblock The {PASCAL} visual object classes (voc) challenge.
\newblock {\em International Journal of Computer Vision}, 88(2):303--338, 2010.

\bibitem{Lin2014}
T.-Y. Lin et~al.
\newblock Microsoft coco: Common objects in context.
\newblock 2014.

\bibitem{63}
A.~Bochkovskiy, C.-Y. Wang, and H.-Y.~M. Liao.
\newblock Yolov4: Optimal speed and accuracy of object detection.
\newblock Apr. 2020.

\bibitem{He}
K.~He, X.~Zhang, S.~Ren, and J.~Sun.
\newblock Deep residual learning for image recognition.
\newblock {\em openaccess.thecvf.com}, 2016.

\bibitem{Ma2019}
Z.~Ma, M.~Li, and Y.~Wang.
\newblock Pan: Path integral based convolution for deep graph neural networks.
\newblock Apr. 2019.

\bibitem{Yao2021}
Z.~Yao, Y.~Cao, S.~Zheng, G.~Huang, and S.~Lin.
\newblock Cross-iteration batch normalization.
\newblock In {\em openaccess.thecvf.com}, 2021.

\bibitem{He2023}
S.~He, R.~Bao, J.~Li, P.~E. Grant, and Y.~Ou.
\newblock Accuracy of segment-anything model (sam) in medical image segmentation tasks.
\newblock Apr. 2023.

\bibitem{Solawetz2020}
J.~Solawetz.
\newblock {YOLOv5 New Version - Improvements and Evaluation}.
\newblock {\em Roboflow Blog}, Jun. 2020.

\bibitem{Wang2022}
C.-Y. Wang, A.~Bochkovskiy, and H.-Y. Liao.
\newblock {YOLOv7: Trainable bag-of-freebies sets new state-of-the-art for real-time object detectors}, Jul. 2022.

\bibitem{Wang2023}
Z.~Wang et~al.
\newblock Mosaic representation learning for self-supervised visual pre-training.
\newblock {\em openreview.net}, Feb. 2023.

\bibitem{Solawetz2022}
J.~Solawetz and et~al.
\newblock What’s new in yolov6?, Jul. 2022.

\bibitem{Xu2022}
X.~Xu, Y.~Jiang, W.~Chen, Y.~Huang, Y.~Zhang, and X.~Sun.
\newblock Damo-yolo: A report on real-time object detection design, Dec. 2022.

\bibitem{Ding2021}
X.~Ding, X.~Zhang, N.~Ma, J.~Han, G.~Ding, and J.~Sun.
\newblock Repvgg: Making vgg-style convnets great again.
\newblock In {\em openaccess.thecvf.com}, 2021.

\bibitem{Huang2016}
G.~Huang, Z.~Liu, and Kilian~Q Weinberger.
\newblock Densely connected convolutional networks.
\newblock 2016.

\bibitem{Jocher2023}
G.~Jocher, A.~Chaurasia, and J.~Qiu.
\newblock Yolo by ultralytics, 2023.
\newblock Accessed: February 30, 2023.

\bibitem{Solawetz2022b}
J.~Solawetz.
\newblock Yolov7 - a breakdown of how it works, Jul. 2022.

\bibitem{Solawetz2023}
J.~Solawetz and et~al.
\newblock What is yolov8? the ultimate guide, Jan. 2023.

\bibitem{Jocher2020}
G.~Jocher and et~al.
\newblock ultralytics/yolov5: v3.0, Aug. 2020.

\bibitem{wang2024yolov9}
Chien-Yao Wang and Hong-Yuan~Mark Liao.
\newblock {YOLOv9}: Learning what you want to learn using programmable gradient information.
\newblock 2024.

\bibitem{wang2024yolov10}
Ao~Wang, Hui Chen, Lihao Liu, Kai Chen, Zijia Lin, Jungong Han, and Guiguang Ding.
\newblock Yolov10: Real-time end-to-end object detection.
\newblock {\em arXiv preprint arXiv:2405.14458}, 2024.

\bibitem{RN1}
Addie Ira~Borja PARICO and Tofael AHAMED.
\newblock An aerial weed detection system for green onion crops using the you only look once (yolov3) deep learning algorithm.
\newblock {\em Engineering in Agriculture, Environment and Food}, 13(2):42--48, 2020.

\bibitem{RN2}
Boyu Ying, Yuancheng Xu, Shuai Zhang, Yinggang Shi, and Li~Liu.
\newblock Weed detection in images of carrot fields based on improved yolo v4.
\newblock {\em Traitement du Signal}, 38(2), 2021.

\bibitem{RN3}
Mads Dyrmann, Anders~Krogh Mortensen, Lars Linneberg, Toke~Thomas Høye, and Kim Bjerge.
\newblock Camera assisted roadside monitoring for invasive alien plant species using deep learning.
\newblock {\em Sensors}, 21(18):6126, 2021.

\bibitem{RN4}
Jiqing Chen, Huabin Wang, Hongdu Zhang, Tian Luo, Depeng Wei, Teng Long, and Zhikui Wang.
\newblock Weed detection in sesame fields using a yolo model with an enhanced attention mechanism and feature fusion.
\newblock {\em Computers and Electronics in Agriculture}, 202:107412, 2022.

\bibitem{RN5}
Qifan Wang, Man Cheng, Shuo Huang, Zhenjiang Cai, Jinlin Zhang, and Hongbo Yuan.
\newblock A deep learning approach incorporating yolo v5 and attention mechanisms for field real-time detection of the invasive weed solanum rostratum dunal seedlings.
\newblock {\em Computers and Electronics in Agriculture}, 199:107194, 2022.

\bibitem{RN6}
Benjamin Costello, Olusegun~O Osunkoya, Juan Sandino, William Marinic, Peter Trotter, Boyang Shi, Felipe Gonzalez, and Kunjithapatham Dhileepan.
\newblock Detection of parthenium weed (parthenium hysterophorus l.) and its growth stages using artificial intelligence.
\newblock {\em Agriculture}, 12(11):1838, 2022.

\bibitem{RN7}
Fengying Dang, Dong Chen, Yuzhen Lu, and Zhaojian Li.
\newblock Yoloweeds: a novel benchmark of yolo object detectors for multi-class weed detection in cotton production systems.
\newblock {\em Computers and Electronics in Agriculture}, 205:107655, 2023.

\bibitem{RN8}
Fernando~J Pérez-Porras, Jorge Torres-Sánchez, Francisca López-Granados, and Francisco~J Mesas-Carrascosa.
\newblock Early and on-ground image-based detection of poppy (papaver rhoeas) in wheat using yolo architectures.
\newblock {\em Weed Science}, 71(1):50--58, 2023.

\bibitem{RN9}
Mino Sportelli, Orly~Enrique Apolo-Apolo, Marco Fontanelli, Christian Frasconi, Michele Raffaelli, Andrea Peruzzi, and Manuel Perez-Ruiz.
\newblock Evaluation of yolo object detectors for weed detection in different turfgrass scenarios.
\newblock {\em Applied Sciences}, 13(14):8502, 2023.

\bibitem{RN10}
Xiaojun Jin, Yanxia Sun, Jun Che, Muthukumar Bagavathiannan, Jialin Yu, and Yong Chen.
\newblock A novel deep learning‐based method for detection of weeds in vegetables.
\newblock {\em Pest Management Science}, 78(5):1861--1869, 2022.

\bibitem{RN24}
Yunong Tian, Guodong Yang, Zhe Wang, Hao Wang, En~Li, and Zize Liang.
\newblock Apple detection during different growth stages in orchards using the improved yolo-v3 model.
\newblock {\em Computers and electronics in agriculture}, 157:417--426, 2019.

\bibitem{RN26}
Shaun~M Sharpe, Arnold~W Schumann, and Nathan~S Boyd.
\newblock Goosegrass detection in strawberry and tomato using a convolutional neural network.
\newblock {\em Scientific Reports}, 10(1):9548, 2020.

\bibitem{RN27}
Mohamad~Haniff Junos, Anis~Salwa Mohd~Khairuddin, Subbiah Thannirmalai, and Mahidzal Dahari.
\newblock An optimized yolo‐based object detection model for crop harvesting system.
\newblock {\em IET Image Processing}, 15(9):2112--2125, 2021.

\bibitem{RN28}
Wenkang Chen, Shenglian Lu, Binghao Liu, Ming Chen, Guo Li, and Tingting Qian.
\newblock Citrusyolo: A algorithm for citrus detection under orchard environment based on yolov4.
\newblock {\em Multimedia Tools and Applications}, 81(22):31363--31389, 2022.

\bibitem{RN29}
Qingqing Hong, Ling Jiang, Zhenghua Zhang, Shu Ji, Chen Gu, Wei Mao, Wenxi Li, Tao Liu, Bin Li, and Changwei Tan.
\newblock A lightweight model for wheat ear fusarium head blight detection based on rgb images.
\newblock {\em Remote Sensing}, 14(14):3481, 2022.

\bibitem{RN30}
Liang Wang, Lingmin Li, Hao Wang, Shaohua Zhu, Zhiqiang Zhai, and Zhongxiang Zhu.
\newblock Real-time vehicle identification and tracking during agricultural master-slave follow-up operation using improved yolo v4 and binocular positioning.
\newblock {\em Proceedings of the Institution of Mechanical Engineers, Part C: Journal of Mechanical Engineering Science}, 237(6):1393--1404, 2023.

\bibitem{RN32}
Juan Wang, Nan Wang, Lihua Li, and Zhenhui Ren.
\newblock Real-time behavior detection and judgment of egg breeders based on yolo v3.
\newblock {\em Neural Computing and Applications}, 32:5471--5481, 2020.

\bibitem{RN33}
Anne~K Schütz, Verena Schöler, E~Tobias Krause, Mareike Fischer, Thomas Müller, Conrad~M Freuling, Franz~J Conraths, Mario Stanke, Timo Homeier-Bachmann, and Hartmut~HK Lentz.
\newblock Application of yolov4 for detection and motion monitoring of red foxes.
\newblock {\em Animals}, 11(6):1723, 2021.

\bibitem{RN35}
Zhenwei Yu, Yuehua Liu, Sufang Yu, Ruixue Wang, Zhanhua Song, Yinfa Yan, Fade Li, Zhonghua Wang, and Fuyang Tian.
\newblock Automatic detection method of dairy cow feeding behaviour based on yolo improved model and edge computing.
\newblock {\em Sensors}, 22(9):3271, 2022.

\bibitem{RN36}
Wael~M Elmessery, Joaquín Gutiérrez, Gomaa~G Abd El-Wahhab, Ibrahim~A Elkhaiat, Ibrahim~S El-Soaly, Sadeq~K Alhag, Laila~A Al-Shuraym, Mohamed~A Akela, Farahat~S Moghanm, and Mohamed~F Abdelshafie.
\newblock Yolo-based model for automatic detection of broiler pathological phenomena through visual and thermal images in intensive poultry houses.
\newblock {\em Agriculture}, 13(8):1527, 2023.

\bibitem{barreiros2021zebrafish}
Marta de~Oliveira Barreiros, Diego de~Oliveira Dantas, Lu{\'\i}s Claudio de~Oliveira Silva, Sidarta Ribeiro, and Allan~Kardec Barros.
\newblock Zebrafish tracking using yolov2 and kalman filter.
\newblock {\em Scientific reports}, 11(1):3219, 2021.

\bibitem{RN37}
Kristina Rančić, Boško Blagojević, Atila Bezdan, Bojana Ivošević, Bojan Tubić, Milica Vranešević, Branislav Pejak, Vladimir Crnojević, and Oskar Marko.
\newblock Animal detection and counting from uav images using convolutional neural networks.
\newblock {\em Drones}, 7(3):179, 2023.

\bibitem{RN38}
Zhiyang Zheng, Jingwen Li, and Lifeng Qin.
\newblock Yolo-byte: An efficient multi-object tracking algorithm for automatic monitoring of dairy cows.
\newblock {\em Computers and Electronics in Agriculture}, 209:107857, 2023.

\bibitem{hao2023yolov5}
Wangli Hao, Li~Zhang, Meng Han, Kai Zhang, Fuzhong Li, Guoqiang Yang, and Zhenyu Liu.
\newblock Yolov5-sa-fc: A novel pig detection and counting method based on shuffle attention and focal complete intersection over union.
\newblock {\em Animals}, 13(20):3201, 2023.

\bibitem{kim2022embeddedpigcount}
Jonggwan Kim, Yooil Suh, Junhee Lee, Heechan Chae, Hanse Ahn, Yongwha Chung, and Daihee Park.
\newblock Embeddedpigcount: Pig counting with video object detection and tracking on an embedded board.
\newblock {\em Sensors}, 22(7):2689, 2022.

\bibitem{RN12}
Jun Liu and Xuewei Wang.
\newblock Tomato diseases and pests detection based on improved yolo v3 convolutional neural network.
\newblock {\em Frontiers in plant science}, 11:898, 2020.

\bibitem{RN13}
Achyut Morbekar, Ashi Parihar, and Rashmi Jadhav.
\newblock Crop disease detection using yolo.
\newblock In {\em 2020 International Conference for Emerging Technology (INCET)}, pages 1--5, 2020.

\bibitem{RN14}
G~Nihar, V~Raghavendra, V~Suresh, and M~Sandhya.
\newblock Rice crop disease detection using yolo algorithm.
\newblock In {\em National Conference On Advances in Electronics Signal Processing and Communications (AESPC-2020)}, volume~6, 2020.

\bibitem{RN15}
Ma~Kristin Agbulos, Yovito Sarmiento, and Jocelyn Villaverde.
\newblock Identification of leaf blast and brown spot diseases on rice leaf with yolo algorithm.
\newblock In {\em 2021 IEEE 7th International Conference on Control Science and Systems Engineering (ICCSSE)}, pages 307--312. IEEE, 2021.

\bibitem{RN16}
Martina Lippi, Niccolò Bonucci, Renzo~Fabrizio Carpio, Mario Contarini, Stefano Speranza, and Andrea Gasparri.
\newblock A yolo-based pest detection system for precision agriculture.
\newblock In {\em 2021 29th Mediterranean Conference on Control and Automation (MED)}, pages 342--347, 2021.

\bibitem{RN17}
Monalika~Padma Reddy and A~Deeksha.
\newblock Mulberry leaf disease detection using yolo.
\newblock {\em International Journal of Advance Research, Ideas and Innovations in Technology}, 7:1816--1821, 2021.

\bibitem{RN18}
Midhun~P Mathew and Therese Yamuna~Mahesh.
\newblock Determining the region of apple leaf affected by disease using yolo v3.
\newblock In {\em 2021 International Conference on Communication, Control and Information Sciences (ICCISc)}, volume~1, pages 1--4, 2021.

\bibitem{RN19}
Shani Verma, Shrivishal Tripathi, Anurag Singh, Muneendra Ojha, and Ravi~R Saxena.
\newblock Insect detection and identification using yolo algorithms on soybean crop.
\newblock In {\em TENCON 2021 - 2021 IEEE Region 10 Conference (TENCON)}, pages 272--277, 2021.

\bibitem{RN20}
Nidhi Kundu, Geeta Rani, and Vijaypal~Singh Dhaka.
\newblock Seeds classification and quality testing using deep learning and yolo v5.
\newblock In {\em Proceedings of the International Conference on Data Science, Machine Learning and Artificial Intelligence}, pages 153--160, 2021.

\bibitem{RN21}
Midhun~P Mathew and Therese~Yamuna Mahesh.
\newblock Leaf-based disease detection in bell pepper plant using yolo v5.
\newblock {\em Signal, Image and Video Processing}, pages 1--7, 2022.

\bibitem{RN22}
Md~Janibul~Alam Soeb, Md~Fahad Jubayer, Tahmina~Akanjee Tarin, Muhammad~Rashed Al~Mamun, Fahim~Mahafuz Ruhad, Aney Parven, Nabisab~Mujawar Mubarak, Soni~Lanka Karri, and Islam~Md Meftaul.
\newblock Tea leaf disease detection and identification based on yolov7 (yolo-t).
\newblock {\em Scientific reports}, 13(1):6078, 2023.

\bibitem{RN23}
Zhenyang Xue, Renjie Xu, Di~Bai, and Haifeng Lin.
\newblock Yolo-tea: A tea disease detection model improved by yolov5.
\newblock {\em Forests}, 14(2):415, 2023.

\bibitem{RN39}
Min Li, Zhijie Zhang, Liping Lei, Xiaofan Wang, and Xudong Guo.
\newblock Agricultural greenhouses detection in high-resolution satellite images based on convolutional neural networks: Comparison of faster r-cnn, yolo v3 and ssd.
\newblock {\em Sensors}, 20(17):4938, 2020.

\bibitem{RN40}
Shahbaz Khan, Muhammad Tufail, Muhammad~Tahir Khan, Zubair~Ahmad Khan, and Shahzad Anwar.
\newblock Deep learning-based identification system of weeds and crops in strawberry and pea fields for a precision agriculture sprayer.
\newblock {\em Precision Agriculture}, 22(6):1711--1727, 2021.

\bibitem{RN41}
Nariman Mamdouh and Ahmed Khattab.
\newblock Yolo-based deep learning framework for olive fruit fly detection and counting.
\newblock {\em IEEE Access}, 9:84252--84262, 2021.

\bibitem{mao2024improved}
Wenjie Mao, Gang Li, and Xiaowei Li.
\newblock Improved re-parameterized convolution for wildlife detection in neighboring regions of southwest china.
\newblock {\em Animals}, 14(8):1152, 2024.

\bibitem{shi2023multi}
Jiayou Shi, Yuhao Bai, Jun Zhou, and Baohua Zhang.
\newblock Multi-crop navigation line extraction based on improved yolo-v8 and threshold-dbscan under complex agricultural environments.
\newblock {\em Agriculture}, 14(1):45, 2023.

\bibitem{chien2024yolov9}
Chun-Tse Chien, Rui-Yang Ju, Kuang-Yi Chou, and Jen-Shiun Chiang.
\newblock Yolov9 for fracture detection in pediatric wrist trauma x-ray images.
\newblock {\em arXiv preprint arXiv:2403.11249}, 2024.

\bibitem{boudaa2024advancing}
Boudjemaa Boudaa, Kamel Abada, Walid~Aymen Aichouche, and Ahmed~Nabil Belakermi.
\newblock Advancing plant diseases detection with pre-trained yolo models.
\newblock In {\em 2024 6th International Conference on Pattern Analysis and Intelligent Systems (PAIS)}, pages 1--6. IEEE, 2024.

\bibitem{kiala2022determining}
Zolo Kiala, John Odindi, and Onisimo Mutanga.
\newblock Determining the capability of the tree-based pipeline optimization tool (tpot) in mapping parthenium weed using multi-date sentinel-2 image data.
\newblock {\em Remote Sensing}, 14(7):1687, 2022.

\bibitem{su2020advanced}
Wen-Hao Su.
\newblock Advanced machine learning in point spectroscopy, rgb-and hyperspectral-imaging for automatic discriminations of crops and weeds: A review.
\newblock {\em Smart Cities}, 3(3):767--792, 2020.

\bibitem{shendryk2020leveraging}
Yuri Shendryk, Natalie~A Rossiter-Rachor, Samantha~A Setterfield, and Shaun~R Levick.
\newblock Leveraging high-resolution satellite imagery and gradient boosting for invasive weed mapping.
\newblock {\em IEEE Journal of Selected Topics in Applied Earth Observations and Remote Sensing}, 13:4443--4450, 2020.

\bibitem{zahid2023lightweight}
Arsalan Zahid, Muhammad Hussain, Richard Hill, and Hussain Al-Aqrabi.
\newblock Lightweight convolutional network for automated photovoltaic defect detection.
\newblock In {\em 2023 9th International Conference on Information Technology Trends (ITT)}, pages 133--138. IEEE, 2023.

\bibitem{hussain2023exudate}
Muhammad Hussain.
\newblock Exudate detection: Integrating retinal-based affine mapping and design flow mechanism to develop lightweight architectures.
\newblock {\em IEEE Access}, 2023.

\bibitem{ahn2021ensemblepigdet}
Hanse Ahn, Seungwook Son, Heegon Kim, Sungju Lee, Yongwha Chung, and Daihee Park.
\newblock Ensemblepigdet: Ensemble deep learning for accurate pig detection.
\newblock {\em Applied Sciences}, 11(12):5577, 2021.

\bibitem{li2023high}
Manzhou Li, Siyu Cheng, Jingyi Cui, Changxiang Li, Zeyu Li, Chang Zhou, and Chunli Lv.
\newblock High-performance plant pest and disease detection based on model ensemble with inception module and cluster algorithm.
\newblock {\em Plants}, 12(1):200, 2023.

\bibitem{lee2022detecting}
Sangyeon Lee, Amarpreet~Singh Arora, and Choa~Mun Yun.
\newblock Detecting strawberry diseases and pest infections in the very early stage with an ensemble deep-learning model.
\newblock {\em Frontiers in Plant Science}, 13:991134, 2022.

\bibitem{singh2023object}
Priya Singh and Rajalakshmi Krishnamurthi.
\newblock Object detection using deep ensemble model for enhancing security towards sustainable agriculture.
\newblock {\em International Journal of Information Technology}, 15(6):3113--3126, 2023.

\bibitem{jin2021automatic}
Rui Jin and Qiang Niu.
\newblock Automatic fabric defect detection based on an improved yolov5.
\newblock {\em Mathematical Problems in Engineering}, 2021, 2021.

\bibitem{zhang2022improved}
Jiaqi Zhang, Junfeng Jing, Pengwen Lu, and Shaojun Song.
\newblock Improved mobilenetv2-ssdlite for automatic fabric defect detection system based on cloud-edge computing.
\newblock {\em Measurement}, 201:111665, 2022.

\bibitem{article-53}
Bing Wei, Bailing Xu, Kuangrong Hao, and Lei Gao.
\newblock Textile defect detection using multilevel and attentional deep learning network (mlma-net).
\newblock {\em Textile Research Journal}, 92:004051752110737, 02 2022.

\bibitem{inproceedings-58}
Zhengrui Peng, Xinyi Gong, Zhenfeng Lu, Xiangyi Xu, Bengang Wei, and Mukesh Prasad.
\newblock A novel fabric defect detection network based on attention mechanism and multi-task fusion.
\newblock In {\em 2021 7th IEEE International Conference on Network Intelligence and Digital Content (IC-NIDC)}, pages 484--488, 11 2021.

\bibitem{article-62}
Zhoufeng Liu, Zhaochen Huo, Chunlei Li, Yan Dong, and Bicao Li.
\newblock Dlse-net: A robust weakly supervised network for fabric defect detection.
\newblock {\em Displays}, 68:102008, 07 2021.

\bibitem{rong2021fabric}
Liu Rong-qiang, Li~Ming-hui, Shi Jia-chen, and Liang Yi-bin.
\newblock Fabric defect detection method based on improved u-net.
\newblock In {\em Journal of Physics: Conference Series}, volume 1948, page 012160. IOP Publishing, 2021.

\bibitem{amigo2019preprocessing}
Jos{\'e}~Manuel Amigo and Carolina Santos.
\newblock Preprocessing of hyperspectral and multispectral images.
\newblock In {\em Data handling in science and technology}, volume~32, pages 37--53. Elsevier, 2019.

\end{thebibliography}

\end{document}